%% file: neurips_2026.tex
\documentclass{article}

 \usepackage[main, final]{neurips_2026}

\usepackage{setspace}                       
\usepackage[utf8]{inputenc} 
\usepackage[T1]{fontenc}    
\usepackage{hyperref}       
\usepackage{url}            
\usepackage{booktabs}       
\usepackage{amsfonts}       
\usepackage{microtype}      

\usepackage{graphicx}
\usepackage{amsmath}
\usepackage{array}     
\usepackage{enumitem}
\usepackage{arydshln}
\usepackage{caption} 
\usepackage{multirow} 
\usepackage{bbm}
\usepackage{lipsum} 
\usepackage{subcaption} 
\usepackage[table]{xcolor}
\usepackage{wrapfig}
\usepackage[ruled,linesnumbered]{algorithm2e}
\usepackage{cleveref}

\usepackage{mathtools} 
\usepackage{bm} 
\usepackage{soul} 
\usepackage{nicefrac}
\usepackage{csquotes}
\usepackage{duckuments}

\crefname{figure}{Fig.}{Figs.}    
\crefname{equation}{Eq.}{Eqs.}
\crefname{table}{Tab.}{Tabs.}
\crefname{algorithm}{Alg.}{Algs.}
\crefname{section}{Sec.}{Secs.}

\Crefname{figure}{Figure}{Figures}
\Crefname{equation}{Equation}{Equations}
\Crefname{table}{Table}{Tables}
\Crefname{algorithm}{Algorithm}{Algorithms}
\title{GDMD: Guiding Distribution Matching Distillation with Gradient-Based Reinforcement Learning}
\newcommand{\sname}{\textbf{GDMD}}

\input{authors}

\begin{document}

\maketitle

\input{section/0_abstract}
\input{section/1_introduction}

\input{section/2_releted}

\input{section/3_method}
\input{section/4_experiment}
\input{section/5_conclusion}
\input{section/6_acknowledgement}
{
\small
\bibliographystyle{plain}
\bibliography{main}
}



\newpage
\input{section/X_appendix}


\end{document}

%% file: authors.tex
\author{%
  \textbf{Linwei Dong}$^{1,2}$\quad
  \textbf{Ruoyu Guo}$^{2}$\quad
  \textbf{Ge Bai}$^{2}$\\
  \textbf{Quan Zheng}$^{3}$\quad
  \textbf{Yawei Luo}$^{1,}$\thanks{Corresponding author.}\quad
  \textbf{Changqing Zou}$^{1,4}$\\[4pt]
  $^1$Zhejiang University\quad
  $^2$ByteDance Inc.\\
  $^3$Institute of Software, CAS \quad
  $^4$Zhejiang Lab\\[4pt]
  \texttt{donglinwei@zju.edu.cn}\quad
  \texttt{yaweiluo@zju.edu.cn}
}

%% file: section/0_abstract.tex
\begin{abstract}

Diffusion distillation, exemplified by Distribution Matching Distillation (DMD), has shown great promise in few-step generation but often sacrifices quality for sampling speed. 
While integrating Reinforcement Learning (RL) into distillation offers potential, a naive fusion of these two objectives relies on suboptimal raw sample evaluation. 
This sample-based scoring creates inherent conflicts with the distillation trajectory and produces unreliable rewards due to the noisy nature of early-stage generation.
To overcome these limitations, we propose \textbf{GDMD}, a novel framework that redefines the reward mechanism by prioritizing distillation gradients over raw pixel outputs as the primary signal for optimization. 
By reinterpreting the DMD gradients as implicit target tensors, our framework enables existing reward models to directly evaluate the quality of distillation updates. 
This gradient-level guidance functions as an adaptive weighting that synchronizes the RL policy with the distillation objective, effectively neutralizing optimization divergence.
Empirical results show that \textbf{GDMD} sets a new SOTA for few-step generation. 
Specifically, our 4-NFE (Number of Function Evaluations‌) RL models outperform the quality of their multi-step teacher and exceed previous DMDR results in GenEval and human-preference metrics, exhibiting strong scalability potential.

\end{abstract}

%% file: section/1_introduction.tex
\input{fig/latar}
\section{Introduction}
\label{sec:intro}
Diffusion models \cite{lipman2022flow, song2020score, saharia2022photorealistic, rombach2022high, ho2020denoising} have revolutionized image synthesis, offering unparalleled realism and training stability compared to GANs \cite{goodfellow2014generative} and VAEs \cite{kingma2013auto}. 
However, their iterative sampling nature remains computationally expensive, requiring numerous network evaluations that hinder real-time interactivity. 
To address this, distillation methods \cite{yin2024one, lu2024simplifying, sauer2024adversarial, song2023consistency, geng2025mean, sauer2024fast, salimans2022progressive} have emerged to compress these models into efficient, few-step generators. 
Yet, these approaches often suffer from quality degradation.
By focusing strictly on minimizing Jensen-Shannon or Kullback-Leibler (KL) divergence to align with the teacher’s distribution, they overlook human aesthetic preferences and real-world data nuances.

Recent advancements have attempted to bridge this quality gap by integrating Reinforcement Learning (RL) \cite{schulman2017proximal, schulman2015trust, black2024trainingdiffusionmodelsreinforcement, wallace2024diffusion} into the Distribution Matching Distillation (DMD) pipeline \cite{black2024trainingdiffusionmodelsreinforcement, jiang2025distribution}. 
Methods like DMDR \cite{jiang2025distribution} pioneer this joint training paradigm, demonstrating RL’s potential to guide distillation towards high-reward regions. 
However, the naive fusion of DMD and RL objectives often triggers optimization instabilities. 
Existing methods \cite{liu2025flowgrpo, xue2025dancegrpo, zheng2025diffusionnft, xu2023imagereward} evaluate the student's raw sample $x_{0}$ independently of the distillation objective. 
This naive sample-based scoring provides suboptimal gradients for few-step distillation, since $x_0$ yields noisy reward feedback in the early stages \cite{jiang2025distribution}.
Thus a preliminary cold-start distillation phase is required to stabilize the joint optimization process.
Furthermore, optimizing an independent RL objective on $x_0$ creates gradient conflicts when few-step learning hits a bottleneck, effectively capping the potential for sample quality of the student model.

To this end, we propose a paradigm shift in DMD-RL distillation: 
leveraging optimization gradients as the primary reward signal, rather than raw sample pixels.
Based on this insight, we propose \sname, a novel framework that treats the gradients derived from the DMD objective as the guiding target for RL-based optimization.
Our carefully designed components cover data collection, gradient evaluation, and policy optimization throughout the RL process, reconstructing the original sample-based scoring into what we call \textbf{gradient-based scoring}.
Our method offers the following advantages over existing methods: 
(1) It provides highly reliable guidance even during early training, bypassing the pitfalls of evaluating noisy initial samples; 
(2) It cleverly leverages the optimization algorithm of DMD, eliminating the need for stochastic differential equation (SDE) sampling to roll out data;
(3) It unlocks higher performance upper bounds via perceptual gradients evaluation and selection.

Extensive experiments demonstrate that GDMD effectively bridges the gap between efficient distillation and human preference alignment.
GDMD achieves state-of-the-art performance in few-step generative modeling, producing models that consistently outperform their multi-step SFT baselines. 
Our comprehensive experiments show that training exclusively on CLIP Score \cite{hessel2021clipscore} and HPS \cite{wu2023human} achieves improvements in other unseen quality metrics (Pick Score \cite{kirstain2023pick}, Aesthetic Score \cite{schuhmann2022laion} and ImageReward \cite{xu2023imagereward}) that surpass existing DMD/DMD-RL-based methods.
Our comparative experiments on the GenEval benchmark further demonstrate the effectiveness of our joint training approach, outperforming competing methods across all evaluation dimensions, as shown in \cref{fig:latar}.

Our key contributions are summarized as follows:
\begin{itemize}
    \item We pioneer a \textbf{gradient-based scoring} strategy for DMD-RL joint distillation and conduct a detailed analysis of its advantages compared with the sample-based scoring strategy.
    \item We design a comprehensive and specialized pipeline, including \textbf{Distillation-aware Gradient Collection (DaGC)}, \textbf{Implicit Gradient Scoring (IGS)}, and \textbf{Negative-aware Preference Optimization (NaPO)}, to effectively harness gradient-level guidance.
    \item We achieve DMD-RL joint training and demonstrate exceptionally high generation quality, as shown in \cref{fig:teaser}.
    We establish a new state-of-the-art for few-step diffusion models, outperforming existing methods on both quality-based (e.g. ImageReward) and rule-based (GenEval) evaluation benchmarks.
\end{itemize}

\input{fig/teaser}

%% file: fig/latar.tex
\begin{figure}[!t]
  \centering
    \includegraphics[width=\linewidth]{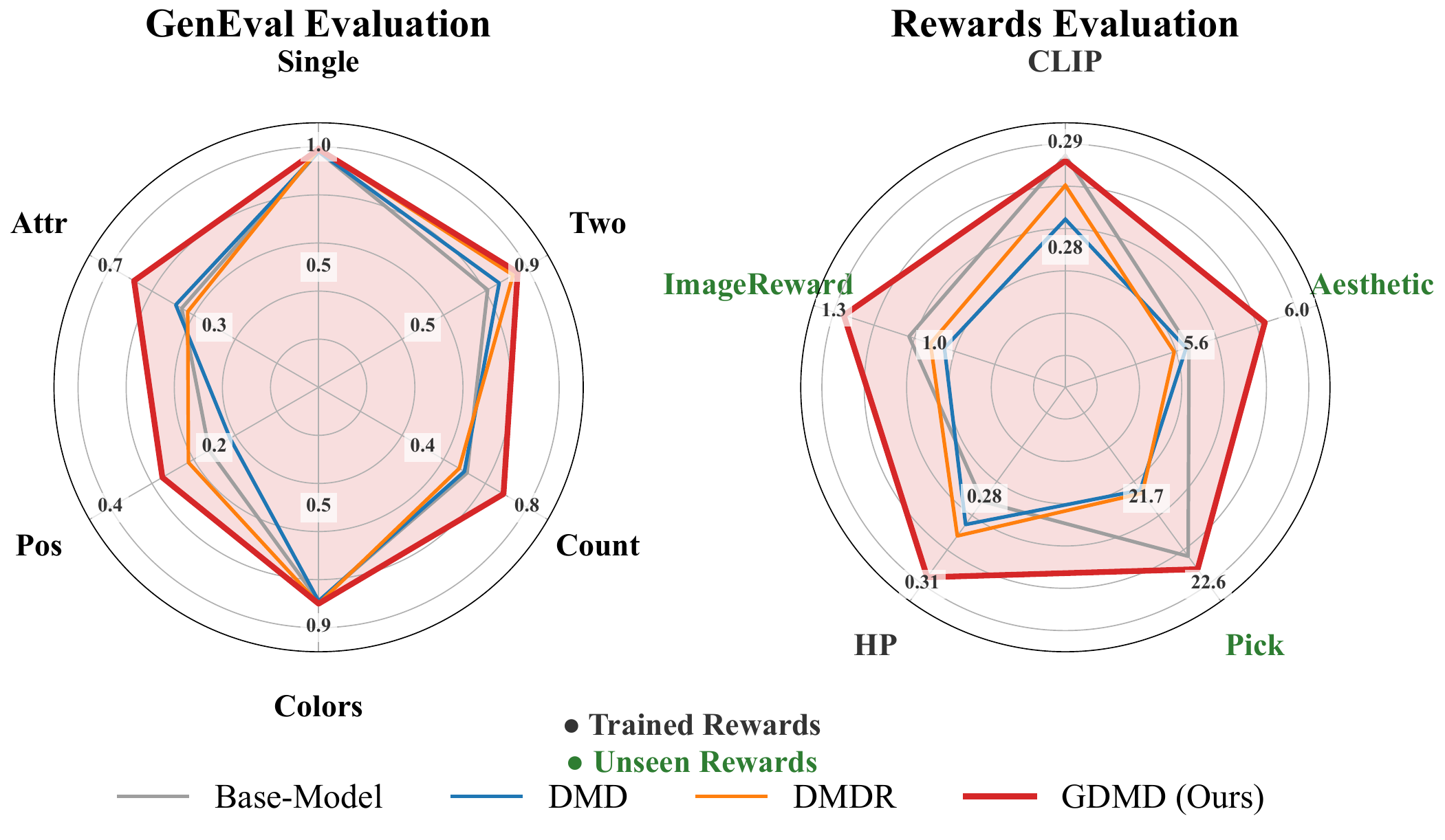}
   \caption{\textbf{Performance of GDMD}. \textbf{Left: } Head-to-head comparison with DMDR on the GenEval benchmark, demonstrating high quality generation. \textbf{Right:} By integrating multiple reward models, GDMD significantly boosts performance across all evaluated benchmarks (including other unseen rewards, such as Aesthetic Score and ImageReward), significantly outperforming existing methods.}
   \label{fig:latar}
\end{figure}

%% file: fig/teaser.tex
\begin{figure}[!t]
  \centering
    \includegraphics[width=\linewidth]{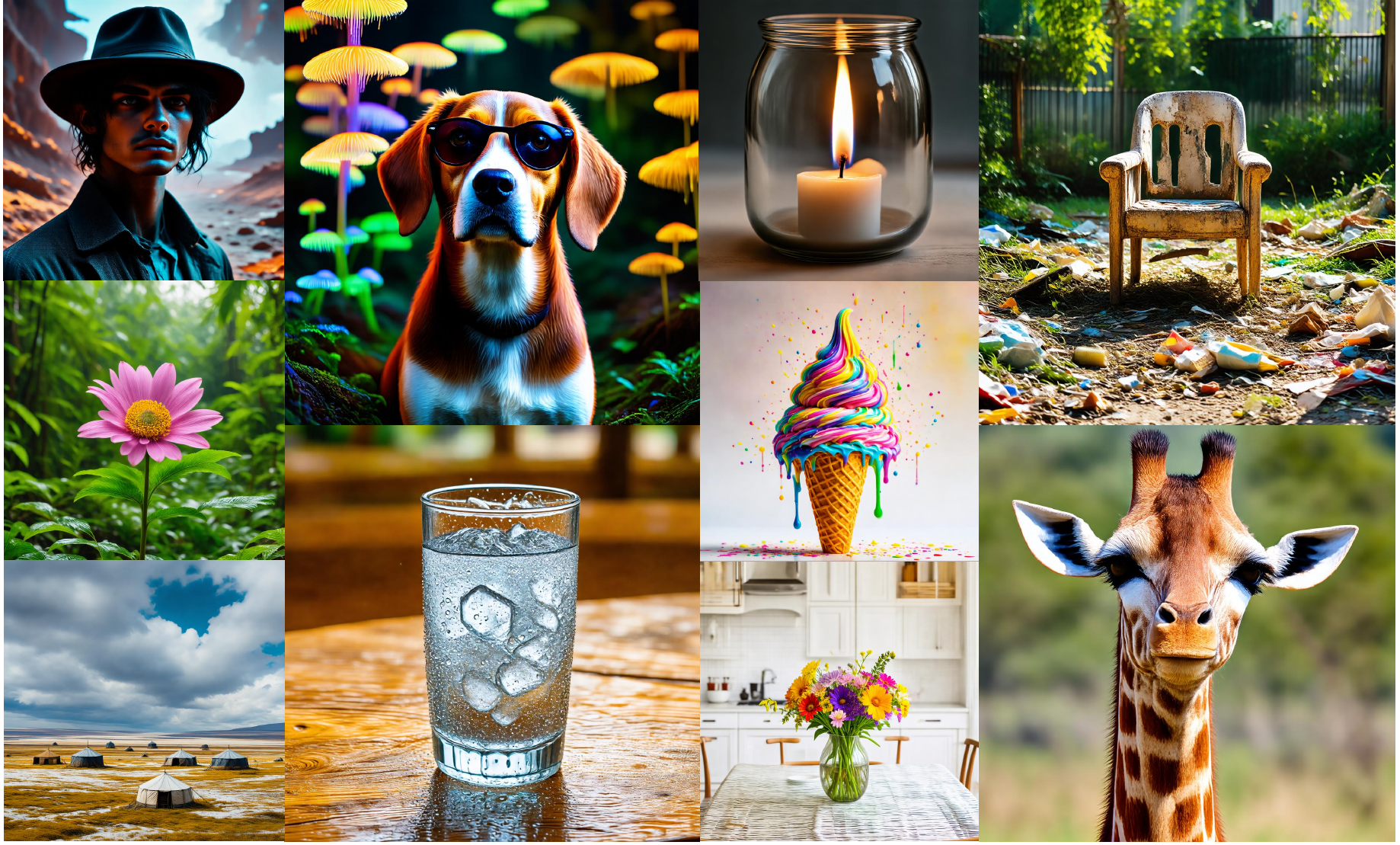}
   \caption{
   \textbf{Samples from 4-NFE student model distilled through our methods.} 
   GDMD showcases outstanding image generation, delivering ultra-realistic visuals and profound concept understanding. 
   }
   \label{fig:teaser}
\end{figure}

%% file: section/2_releted.tex
\section{Related Works}
\label{sec:related}
\subsection{Diffusion Distillation}
Diffusion distillation transfers a pretrained model's generative capability to a one- or few-step student through trajectory matching, distribution matching, or combinations with adversarial learning.

\textbf{Trajectory-based methods} follow the teacher's sampling path in fewer steps.
Progressive Distillation \cite{salimans2022progressive} compresses two teacher steps into one student step; 
Consistency Models \cite{song2023consistency} map points on the same ODE trajectory to a common endpoint, with LCM \cite{luo2023latent} doing so in latent space; CTM \cite{kim2024consistency} extends consistency to mappings between trajectory states. 
More recent continuous-time formulations \cite{lu2024simplifying, geng2025mean, dong2026dual,frans2025one} further refine this trajectory-matching perspective toward efficient one-step generation.

\textbf{Score- and distribution-based methods} use teacher score information to guide the student's output distribution without requiring pointwise reproduction of teacher trajectories. 
Score distillation, initially popularized in text-to-3D generation \cite{poole2022dreamfusion,wang2023prolificdreamer}, also underlies diffusion-generator training in Diff-Instruct \cite{luo2023diff} and SiD \cite{zhou2024score}. 
DMD \cite{yin2024one} matches diffused student and teacher distributions using score differences, and subsequent work improves its training and guidance \cite{yin2024improved,liu2025decoupled,luo2025learning}. 

\textbf{Adversarial and hybrid methods} combine complementary supervision: 
ADD \cite{sauer2024adversarial} and LADD \cite{sauer2024fast} incorporate adversarial learning into diffusion distillation, 
SDXL-Lightning \cite{lin2024sdxl} combines progressive and adversarial distillation, 
and DMD2 \cite{yin2024improved} adds adversarial supervision to distribution matching. 
These approaches overlap rather than form disjoint categories. 
Beyond adversarial supervision, DMDR \cite{jiang2025distribution} combines DMD with reward alignment. 
Our work follows this joint-training direction and evaluates DMD-induced update targets to guide preference optimization.

\subsection{Reinforcement Learning in Diffusion}
Early efforts \cite{schulman2015trust, schulman2017proximal} to align diffusion models with human preferences predominantly relied on conventional Reinforcement Learning (RL) frameworks. 
Drawing inspiration from Proximal Policy Optimization (PPO) \cite{schulman2017proximal}, several foundational works \cite{black2024trainingdiffusionmodelsreinforcement, fan2023dpokreinforcementlearningfinetuning, lee2023aligningtexttoimagemodelsusing, wallace2024diffusion} incorporate policy gradient techniques by optimizing the score function \cite{song2020score}.
More recently, a significant shift has emerged with the introduction of Group Relative Policy Optimization (GRPO) \cite{shao2024deepseekmath}. 
Unlike traditional RL methods that depend on a separately trained value network, GRPO leverages relative rewards computed within a sampled group to reduce variance without requiring a critic model. 
Building on this, FlowGRPO \cite{liu2025flowgrpo} and DanceGRPO \cite{xue2025dancegrpo} extend GRPO-style updates to flow-matching models by converting deterministic ordinary differential equation (ODE) sampling into stochastic differential equation (SDE) formulations. 
Subsequent works \cite{li2025branchgrpo, he2025gardo, li2026mixgrpounlockingflowbasedgrpo, wang2025grpo, he2025tempflow} have further refined GRPO-based frameworks to enhance both training efficiency and stability.
Despite these advances, recent studies \cite{li2025uniworld, xue2025advantage, zheng2025diffusionnft} highlight key limitations of likelihood-based policy optimization, including systematic bias and restricted solver flexibility.
In response, DiffusionNFT \cite{zheng2025diffusionnft} proposes a novel approach that integrates reinforcement signals directly into the standard diffusion training objective, bypassing the need for explicit likelihood estimation or SDE-based reverse processes.

%% file: section/3_method.tex
\section{Method}
\input{fig/distribution}
\input{section/2_preliminary}

\input{fig/pipeline}
\subsection{Distribution Matching Distillation with On-Policy RL}
Rather than incorporating GANs, DMDR \cite{jiang2025distribution} introduces a joint training framework (illustrated in \cref{fig:pipeline} top) that integrates distillation and reinforcement learning within diffusion models.
However, this approach simply concatenates the losses from the two tasks, thereby overlooking the potential interplay between distillation and reinforcement learning.
It inevitably encounters the following limitations during training:
First, DMDR is heavily contingent on cold starts.
Since the underlying pre-trained model lacks strong few-step generation capabilities, the initial noisy and unstable generation results (See \cref{fig:distribution} (c) top)  may cause evaluation bias in the reward model.
DMDR introduces a cold start stage to address this challenge, incorporating pre-designed iterations that rely solely on DMD distillation before the RL loss is introduced.
This design splits the entire process into two stages, which does not fully leverage the guiding role of RL.
Second, the RL gradient direction may diverge from the DMD gradient direction, as shown in \cref{fig:distribution} (a).
Since DMD fits the teacher model distribution, its gradient loss does not specify the direction of user preferences.
This conflict can lead to misguided updates that converge on a reward hacking state, which ultimately compromises the few-step distillation process.
These inherent limitations render DMDR both inefficient and unstable.

Inspired by the real-fake score estimation in DMD, we evaluate the gradient of DMD rather than the sample $x_0$.
Contrary to the prevailing belief that RL scoring should be conducted on generated samples, our gradient-based scoring strategy offers the following advantages:
\textbf{1. Gradient-based scoring ensures training stability.} 
In the early stages of training, evaluating and optimizing based on gradients is more conducive to stability. 
Since gradient-based scoring does not alter the original gradient descent direction, it only assigns adaptive weights based on gradient quality.
This means we do not require additional cold-start overhead.
In contrast, sample-based scoring evaluates the current state that perturbs the DMD gradient, see \cref{fig:distribution} (b).
\textbf{2. No need for additional SDE designs.}
Unlike traditional data collection \cite{liu2025flowgrpo, xue2025dancegrpo}, gradient information can be obtained by computing score differences across different fake score versions or noise intensities (see \cref{subsec:dagc}). 
This eliminates the need for large amounts of SDE sampling process.
\textbf{3. RL optimization aligns with the distillation direction. }
The gradients evaluated by RL are consistent with the distillation direction of the original DMD. 
This alignment ensures that the evaluation does not interfere with the original few-step distillation process.
\cref{fig:distribution} (b) shows the visualization of the two methods without cold start.

\subsection{Guiding DMD with Gradient-Based RL}
We provide \sname\  pipeline in \cref{fig:pipeline}. 
Below, we elaborate on key design choices.

\paragraph{\bf Distillation-aware Gradient Collection.}
\label{subsec:dagc}
Unlike the SDE sampling employed in Flow-GRPO \cite{liu2025flowgrpo} and DanceGRPO \cite{xue2025dancegrpo}, we utilize the inherent randomness of DMD for grouped gradient collection.
We employ two distinct collection strategies, leveraging  the characteristics of distribution fitting and the dynamic updates of the fake score estimator, respectively.
As shown in \cref{fig:pipeline} below,
$t$-based collection strategy generates diverse gradient directions by estimating scores with varying noise intensities, which benefits fitting the $t \sim \mathcal{U}(0,1000)$ distribution.
$\mu_\text{fake}^\phi$-based collection strategy aims to gather gradients throughout the optimization process by leveraging different versions of the fake score function from updates.
It traces the different iterations of the fake score function, with the RL process favoring gradients produced by more suitable versions.

\paragraph{\bf Implicit Gradient Scoring.}
Traditional reward models, such as CLIP Score \cite{hessel2021clipscore}, HPS \cite{wu2023human}, and ImageReward \cite{xu2023imagereward}, require image and text inputs instead of gradients.
We need to transform or implicitly evaluate the gradient to make it compatible with the existing evaluation function.
Since $ \mathbb{E}[y \cdot \nabla_{\theta}G_{\theta}^{\top} ] = \nabla_{\theta} \mathbb{E}[y \cdot G_{\theta}^{\top} ] = \frac{1}{2} \nabla_{\theta} \mathbb{E}[\|G_{\theta} - (G_{\theta^{-}} - y)\|_2^2]$, \cref{eq:dmd-gen} can be transformed into
\begin{align}
\frac{1}{2} \nabla_\theta \mathbb{E}_{t, z} \big|\big|G_\theta(z) - \big(G_{\theta^{-}}(z) +  ( s_{\text{real}}(\mathbf{F}(x_0, t), t) - s_{\text{fake}}(\mathbf{F}(x_0, t), t) ) \big) \big|\big|^2_2,
\label{eq:gradient-scoring}
\end{align}
where $\theta^{-}$ denotes the stop gradient.
\cref{eq:gradient-scoring} embodies the loss objective $x_{tar}:=G_{\theta^{-}}(z) +  ( s_{\text{real}}(\mathbf{F}(x_0, t), t) - s_{\text{fake}}(\mathbf{F}(x_0, t), t) $ for DMD, which is computed during training using the generated sample $x_0$ and the score estimation.
Intuitively, $x_{tar}$ implicitly contains information about the quality of gradients, as it serves as the final loss objective for DMD.
More detailed analysis is provided in the Appendix \cref{sec:proof}.
Therefore, Given the reward model $R_{\psi}$ and the VAE decoder $\mathcal{D}_\text{vae}$, the final gradient's score $r^{raw}$ for $x_\text{tar}$ and the paired text prompts $c$ can be expressed as
\begin{align}
r^{\text{raw}}(x_\text{tar},c) = R_{\psi}(\mathcal{D}_\text{vae}(x_\text{tar}), c).
\label{eq:rraw}
\end{align}
The value of $r^\text{raw}$ in \cref{eq:rraw} implicitly reflects the quality of the gradients during training, as the equivalence transformation performed in \cref{eq:gradient-scoring} concentrates the gradient information into $x_\text{tar}$.

\paragraph{\bf Negative-aware Preference Optimization.}
Inspired by DiffusionNFT \cite{zheng2025diffusionnft}, we employ implicit positive and negative policies for preference learning.
Motivated by existing practices \cite{liu2025flowgrpo, xue2025dancegrpo}, we first transform the raw reward $r^\text{raw}$ into $r \in [0, 1]$ that represents the optimality probability:
\begin{align}
r(x_\text{tar}, c) := \frac{1}{2} + \frac{1}{2} \text{clamp} \left[ \frac{r^{\text{raw}}(x_\text{tar}, c) - r^{\text{raw}}(x_0, c)}{Z_c}, -1, 1 \right].
\label{eq:norm}
\end{align}
${Z_c > 0}$ serves as a normalizing factor, which could represent the global standard deviation of the reward.
The current sample supplies a prompt-matched baseline: $r>1/2$ favors the candidate target, whereas $r<1/2$ disfavors it. 
Following DiffusionNFT, our RL training objective can be written as:
\begin{align}
\mathcal{L}_\text{RL}^{\theta} = \mathbb{E}_{c, t} \left[ r \|v_\theta^+ (x_t, c, t) - v\|_2^2 + (1 - r) \|v_\theta^- (x_t, c, t) - v\|_2^2 \right],
\label{eq:nft}
\end{align}
where \( v_\theta^+ (x_t, c, t) := (1 - \beta) v^{\text{old}}(x_t, c, t) + \beta v_\theta(x_t, c, t) \), (implicit positive policy)
and \( v_\theta^- (x_t, c, t) := (1 + \beta) v^{\text{old}}(x_t, c, t) - \beta v_\theta(x_t, c, t) \), (implicit negative policy).
$v_\theta(x_t, c, t)$ and $v^{\text{old}}$ represent the velocity fields predicted by different versions of the student model.
$\beta$ is a hyperparameter, usually set to 1, and $v$ is the target velocity field to be aligned, which can be obtained from $x_\text{tar}$.
\cref{eq:nft} presents the $v$-loss for DiffusionNFT policy, which can be converted into an objective containing $x_\text{tar}$ via the solver.
\cref{fig:ap:mode} in the appendix illustrates the optimization of the DiffusionNFT policy.

\paragraph{\bf Training Scheme.}
Here we organize the entire training process. Our final joint training loss for generator $G_\theta$ can be expressed as:
\begin{equation}
    \mathcal{L}_\text{GDMD}^{\theta} = \lambda \mathcal{L}_\text{DMD}^{\theta} + \gamma \mathcal{L}_\text{RL}^{\theta},
    \label{eq:total}
\end{equation}
where $\mathcal{L}_\text{DMD}^{\theta}$ is defined in \cref{eq:dmd-gen}, and
$\mathcal{L}_\text{RL}^{\theta}$ is defined in \cref{eq:nft}.
$\lambda$ and $\gamma$ are the loss adjustment weights.
We use $\lambda=1$ and $\gamma=0.1$; \cref{sec:ap:ablation} studies the balance between distillation and reward alignment.
For updating the fake score function, we use the loss defined in \cref{eq:dmd-fake}.
The entire training algorithm is provided in \cref{alg:distillation}. 
Further implementation details are provided in the Appendix \cref{sec:ap:imple}.

%% file: fig/distribution.tex
\begin{figure}[!t]
  \centering
    \includegraphics[width=\linewidth]{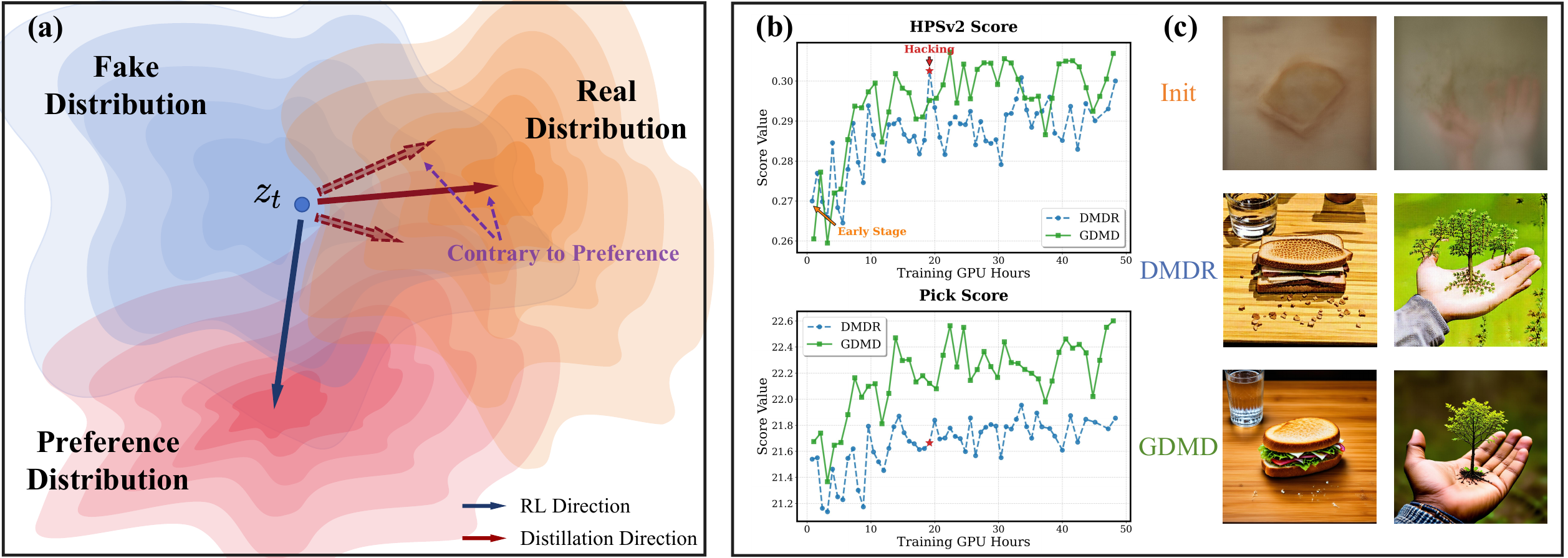}
   \caption{\textbf{(a) Illustration of divergence between RL updates and DMD updates.} 
   RL optimization diverges from few-step distillation in certain extreme phases.
  \textbf{(b) Compared to DMDR without cold start.} 
  DMDR is more prone to reward hacking (background artifacts), and unseen metrics (such as Pick Score) do not show significant improvement.
  \textbf{(c)} \textbf{Visualization of the initial phase and visualization of reward hacking in DMDR.} GDMD has not exhibited any significant hacking.
   }
   \label{fig:distribution}
\end{figure}

%% file: section/2_preliminary.tex
\subsection{Preliminary}
{\bf Distribution Matching Distillation (DMD) \cite{yin2024one}} is proposed to condense a multi-step diffusion model into a one/few-step generator $G_\theta$ by minimizing a time-averaged Kullback-Leibler (KL) divergence between diffused real distribution $p_\text{real}$ and fake distribution $p_\text{fake}$. 
The gradient of this loss reduces to the difference of two score functions:
\begin{align}
\nabla_\theta \mathcal{L}_{\text{DMD}} = - \mathbb{E}_{t, z} \left[ \big( s_{\text{real}}(\mathbf{F}(x_0, t), t) - s_{\text{fake}}(\mathbf{F}(x_0, t), t) \big) \frac{d G_\theta(z)}{d\theta} \right],
\label{eq:dmd-gen}
\end{align}
where $z \sim \mathcal{N} (0, I)$, $x_0 := stopgrad(G_\theta(z))$ , $\mathbf{F}$ applies forward diffusion at time $t$, and $s_{\text{real}}$, $s_{\text{fake}}$ are scores estimated by diffusion models $\mu_{\text{real}}$ and $\mu_{\text{fake}}^{\phi}$. 
During training, $\mu_{\text{real}}$ is a frozen pre-trained teacher, while $\mu_{\text{fake}}^{\phi}$ is updated jointly with $G_\theta$ via denoising matching on generated samples $x_0$:
\begin{align}
\mathcal{L}_{\text{denoise}}^{\phi} = \|\mathcal{S}(\mu_{\text{fake}}^{\phi}(\mathbf{F}(x_0, t), t)) - x_0\|^2_2,
\label{eq:dmd-fake}
\end{align}
where $\mathcal{S}$ is the solver that maps the diffusion outputs to $x_0$.
However, constrained by multi-step teacher models, the upper limit of distilled student models remains capped. 
Pioneering explorations, such as DMD2 \cite{yin2024improved}, have broken through this ceiling by incorporating adversarial training with GAN-style designs.
Nevertheless, these methods often require large amounts of high-quality real data and demand more meticulous parameter tuning to ensure training stability, which introduces potential pitfalls throughout the distillation process.

%% file: fig/pipeline.tex
\begin{figure}[!t]
  \centering
    \includegraphics[width=\linewidth]{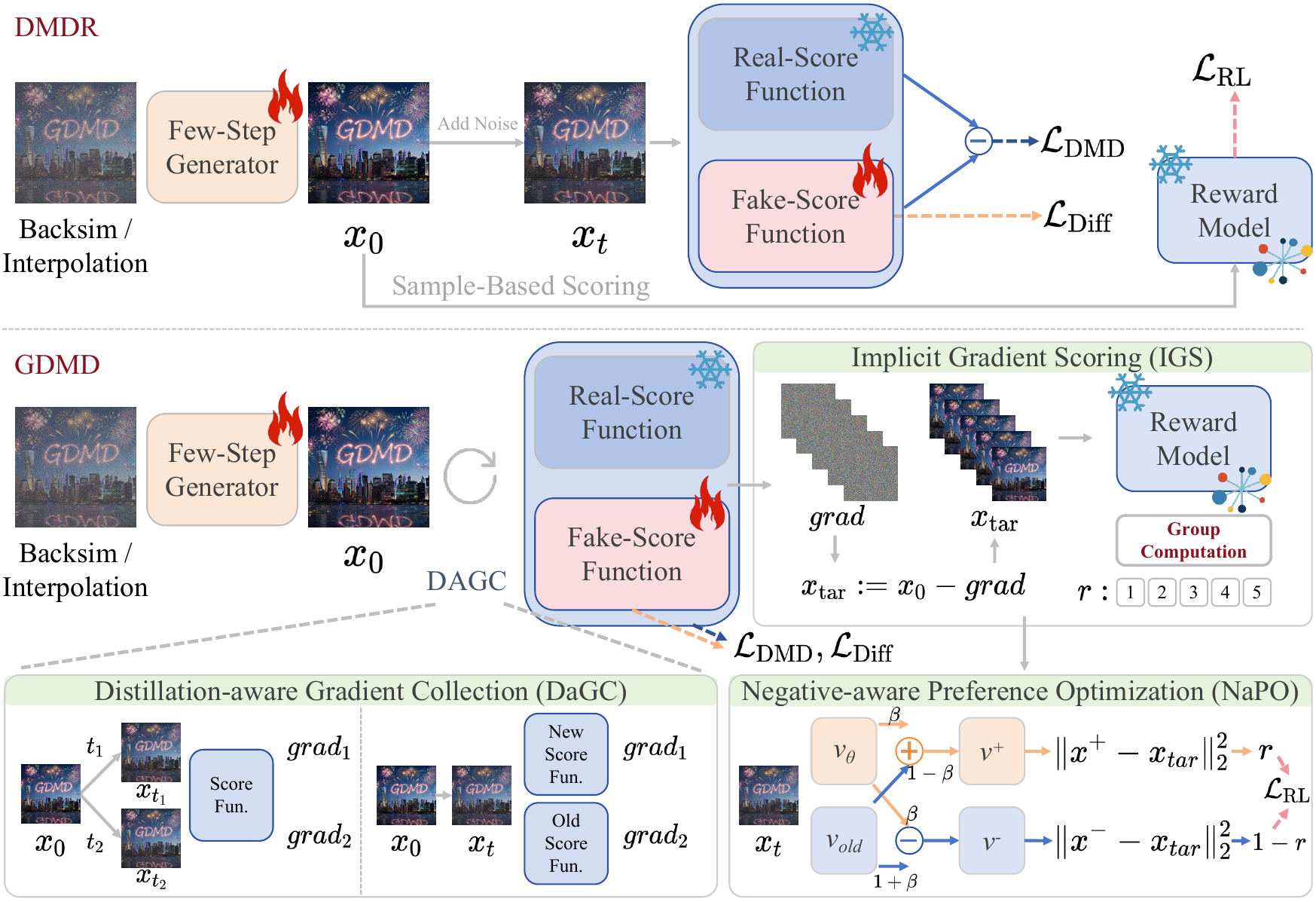}
   \caption{ \textbf{Overview of the DMDR and GDMD pipelines.}
   \textbf{Top:} DMDR employs a sample-based scoring approach, simply combining the losses from DMD and RL.
       \textbf{Bottom:} GDMD achieves gradient-based RL optimization through its carefully designed components: 
       \textbf{Distillation-aware Gradient Collection (DaGC)}, 
       \textbf{Implicit Gradient Scoring (IGS)}, 
       and \textbf{Negative-aware Preference Optimization (NaPO)}, which collectively unlock the model's full performance potential.
   }
   \label{fig:pipeline}
\end{figure}

%% file: section/4_experiment.tex
\section{Experiments}
\input{fig/main_vis}
\label{sec:exp}
\subsection{Experimental Setup}
\label{sec:exp_setup}
We provide a concise description of our training and evaluation setups.
For training, we adopt a distillation approach utilizing the Unet-based model (SDXL-Base \cite{podell2023sdxl}) and the DiT-based model (SD3-Medium \cite{esser2024scaling}), with prompts drawn from the text-to-image-2M dataset \cite{jackyhatetexttoimag2M}.
To assess the performance of GDMD, we conduct extensive experiments. 
Specifically, following prior work \cite{jiang2025distribution}, we generate images using prompts randomly sampled from the high-quality ShareGPT-4o-Image dataset \cite{chen2025sharegpt} and report several evaluation metrics including CLIP Score \cite{hessel2021clipscore}, Aesthetic Score \cite{schuhmann2022laion}, Pick Score \cite{kirstain2023pick}, Human Preference (HP) Score \cite{wu2023human}, and ImageReward \cite{xu2023imagereward}.
For a more holistic evaluation, we also compare our distilled models with their teachers, as well as with DMD \cite{yin2024one} and DMDR \cite{jiang2025distribution}, on the GenEval \cite{ghosh2023geneval} benchmark. 
In addition, we perform both qualitative and quantitative ablation studies to further validate the effectiveness.

\subsection{Qualitative and Quantitative Comparison}
We present a detailed comparison of our method's performance against competing approaches in \cref{fig:main_vis}.
Text misalignment persists in existing methods (e.g., Hyper-SD \cite{ren2024hyper}, Flash Diffusion \cite{chadebec2025flash}, and DMD2 \cite{yin2024improved} do not generate galaxy beards in case 1; Flash Diffusion is missing the blue car, while both DMD2 and DMDR exhibit hallucination by generating another object in case 4.).
For image quality, they also exhibit more distortion issues, such as fingers distortion, body distortion, etc.
Meanwhile, DMDR \cite{jiang2025distribution} and Hyper-SD exhibit stronger oversaturation and glare.
GDMD demonstrates superior performance compared to other methods, exhibiting significant advantages in addressing distortion, saturation, and text alignment.
\cref{fig:teaser} displays our visualization results. 
Our model maintains high-quality images that align with user preferences even under extensive rollouts.

\Cref{tab:system_compare} presents the evaluation results for various image quality metrics following training with CLIP Score and HP Score. 
The experimental results demonstrate that our joint training method not only outperforms the baseline approach, DMDR, but also surpasses all existing distillation methods (e.g. LCM \cite{luo2023latent}, SDXL-Lighting \cite{lin2024sdxl}, DMD \cite{yin2024one} and DMD2 \cite{yin2024improved} for SDXL-Base; Hyper-SD \cite{ren2024hyper}, Flash Diffusion \cite{chadebec2025flash}, TDM \cite{luo2025learning}, DMD and DMD2 for SD3-Base) across both trained (Clip Score, HP Score) and unseen (Aesthetic Scores, Pick Score, and ImageReward) quality metrics.

\Cref{tab:geneval} compares our method with naive DMD and the competitive approach DMDR under GenEval benchmark. 
In distillation settings, our method consistently surpasses the teacher model (under CFG inference) across all evaluation dimensions. 
It also achieves substantial improvements over both DMD and DMDR in terms of generation quality (See \cref{fig:ab_vis}) and GenEval scores, underscoring the effectiveness of reinforcement learning strategies.

\input{tables/main_table}

\input{tables/main_geneval}

\subsection{Human Evaluation}
\input{fig/user-study}
\label{sec:exp_human}
\Cref{fig:user-study} presents the results of our user study comparing our method with DMD, DMDR, DMD+DiffusionNFT (sample-based scoring) and the Base Model. 
The study findings demonstrate that GDMD consistently wins against all compared baselines.
It is preferred by 55.1\% of users over the Teacher model, 61.9\% over DMD + DiffusionNFT, 62.7\% over DMDR and achieves its highest win rate of 64.6\% against DMD in terms of image quality.
For prompt alignment, the win rates are 57.4\%, 57.6\%, 56.0\%, and 61.6\% respectively, demonstrating higher performance.

\subsection{Ablation Study}
\Cref{tab:ab_rl} compares the impact of different RL strategies under the same configuration.
We observe that both ReFL, GRPO and DiffusionNFT methods contribute to score improvements across trained metrics.
However, our approach not only improves the target reward metric but also generalizes effectively to other unseen quality evaluations, demonstrating a genuine and robust enhancement in model performance.
\cref{fig:ab_vis} shows the visualization results of our approach compared to the best-performing competitive method (DMDR with DiffusionNFT policy). 
Our approach achieves superior text alignment, superior image quality, and reduced distortion compared to the baseline. 
\input{fig/ab_vis}
\input{tables/ab_rl}

We present our ablation study on different collection strategies in \cref{tab:ab_trick}.
Experimental results demonstrate that all three strategies contribute to RL training.
Specifically, we found that employing a $t$-based strategy significantly improves training scores, while the $\mu_\text{fake}^\phi$-based strategy demonstrates the best generalization on unseen metrics. 
Using both strategies simultaneously enables the model to explore more effectively, yielding superior performance.
\input{tables/ap/ab_trick}

Further studies are provided in the Appendix \cref{sec:ap:ablation}.

%% file: fig/main_vis.tex
\begin{figure}[!t]
  \centering
    \includegraphics[width=\linewidth]{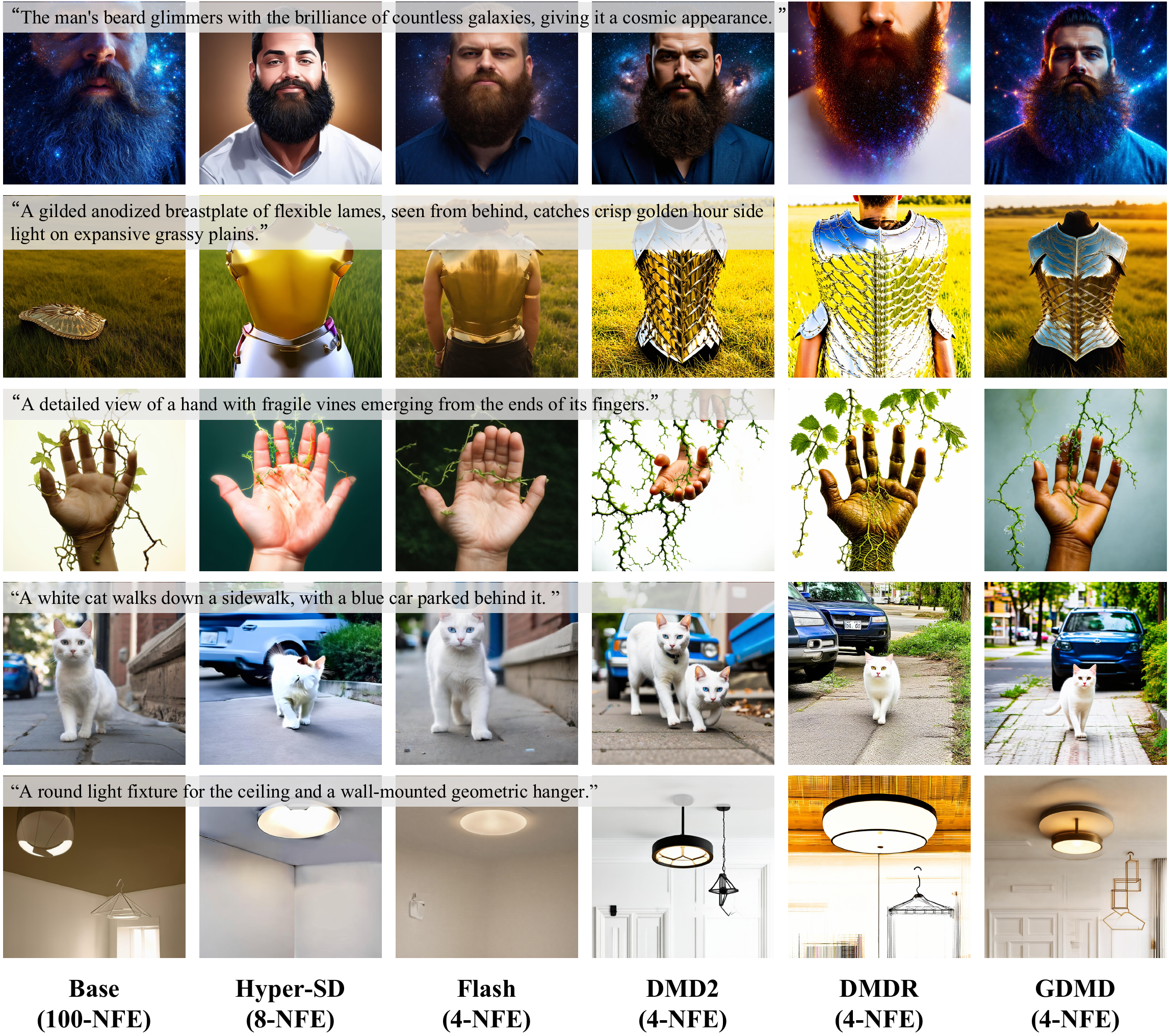}
   \caption{\textbf{Qualitative comparison against teacher and existing models.} Using identical noise inputs, 
   our method outperforms others in both quality and prompt alignment, showing strong performance.
   }
   \label{fig:main_vis}
\end{figure}

%% file: tables/main_table.tex
\begin{table*}[t]
\centering
\captionof{table}{\textbf{Quantitative comparison with state-of-the-art approaches.} $\star$ marks our reproduced results, with the best scores in \textbf{bold}. All RL methods are trained solely on Clip Score and HP Score.}
\vspace{-0.2em}
\resizebox{1.0\linewidth}{!}{
\begin{tabular}{l c c c c c c c}
\toprule
\multirow{2}{*}{\textbf{Method}} & \multirow{2}{*}{\textbf{~Step}} & \multirow{2}{*}{\textbf{~NFE}} & \multicolumn{2}{c}{\textbf{Trained Rewards}} & \multicolumn{3}{c}{\textbf{Unseen Rewards}} \\
\cmidrule(lr){4-5} \cmidrule(lr){6-8}
& & & ~CLIP Score$\uparrow$ & ~HP Score$\uparrow$ & ~Aesthetic Score$\uparrow$ &~Pick Score$\uparrow$ & ~ImageReward$\uparrow$\\
\arrayrulecolor{black}\midrule

\multicolumn{8}{l}{\textcolor{gray}{\emph{SDXL-Base}}} \\
Base-Model \cite{podell2023sdxl} (w/o CFG)& 50 & 50 & 0.2294 & 0.1903 & 5.5312 & 20.5192 & -0.8276 \\
Base-Model (w/ CFG)& 50 & 100 & 0.2872 & 0.2710 & 5.8276 & 22.2560 &  0.7303 \\
\arrayrulecolor{black!40}\midrule
LCM~\cite{luo2023latent} (w/ CFG) & 4 & 8 & 0.2807 & 0.2615 & 5.3214 & 21.9570 & 0.5573 \\
SDXL-Lightning~\cite{lin2024sdxl} & 4 & 4 & 0.2827 & 0.2791 & 5.6128 & 22.3990 & 0.6651 \\
DMD$^{\star}$~\cite{yin2024one} & 4 & 4 & 0.2868 & 0.2732 & 5.7155 & 22.0766 & 0.7640 \\
DMD2~\cite{yin2024improved} & 4 & 4 & 0.2905 & 0.2940 & 5.7391 & 22.4406 & 0.9094 \\
DMDR$^{\star}$~ \cite{jiang2025distribution} & 4 & 4 & 0.2899 & 0.2967 & 5.7912 & 22.2066 & 0.7581\\
\rowcolor[RGB]{240,230,245}
GDMD (Ours) & 4 & 4 & \textbf{0.2912} & \textbf{0.2991} & \textbf{5.8120} & \textbf{22.5446} & \textbf{0.9171}\\

\arrayrulecolor{black!40}\midrule

\multicolumn{8}{l}{\textcolor{gray}{\emph{SD3-Medium}}} \\
Base-Model \cite{esser2024scaling} (w/o CFG) & 50 & 50 & 0.2535 & 0.2010 & 5.2840 & 20.5660 & -0.3762  \\
Base-Model (w/ CFG) & 50 & 100 & 0.2936  & 0.2810 & 5.5711 & 22.3236 & 1.0759 \\
\arrayrulecolor{black!40}\midrule
DMD$^{\star}$~\cite{yin2024one} & 4 & 4 & 0.2861 & 0.2891 & 5.5598 & 21.6216 & 0.9704 \\
DMD2$^{\star}$~\cite{yin2024improved} & 4 & 4 & 0.2914  & 0.2951 &  5.7704 &  22.1442 & 1.1689 \\
Flash Diffusion~\cite{chadebec2025flash} & 4 & 4 & 0.2864 & 0.2636 & 5.5236 &  22.0558 &  0.8752  \\
Hyper-SD~\cite{ren2024hyper} (w/ CFG) & 4 & 8 & 0.2827 & 0.2608 & 5.5715 & 21.2576 & 0.6562  \\
TDM~\cite{luo2025learning} & 4 & 4 & 0.2848 & 0.2940 & 5.7070 & 22.1806 & 1.0932 \\
DMDR$^{\star}$ \cite{jiang2025distribution} & 4 & 4 & {0.2901} & {0.2931} &  {5.5123} &  {21.6528} & {1.0120} \\
\rowcolor[RGB]{240,230,245}  
GDMD (Ours) & 4 & 4 & \textbf{0.2930} & \textbf{0.3076} &  \textbf{5.8728} &  \textbf{22.4614} & \textbf{1.2702} \\

\arrayrulecolor{black}\bottomrule
\end{tabular}}
\vspace{-0.2em}
\label{tab:system_compare}
\end{table*}

%% file: tables/main_geneval.tex
\begin{table*}[t]
\centering
\captionof{table}{\textbf{Quantitative comparison on GenEval} of our GDMD against multi-step teacher, DMD and DMDR. The best result is highlighted in \textbf{bold}.}
\vspace{-0.2em}
\resizebox{1.0\linewidth}{!}{
\begin{tabular}{l c c c c c c c c c}
\toprule
Model & ~~Step & ~~NFE & ~~Overall & ~~Single. & ~~Two. & ~~Count. & ~~Colors & ~~Pos. & ~~Attr. \\
\arrayrulecolor{black}\midrule

Base-Model (w/o CFG) & 50 & 50 &  0.22  &  0.58 & 0.13  & 0.11 & 0.41 & 0.03 & 0.07 \\
Base-Model (w/ CFG) & 50 & 100 & 0.62  & 0.98  & 0.73  & 0.57  & 0.80 & 0.21 & 0.46 \\
DMD & 4 & 4 & 0.63 & 0.98 &  0.78 & 0.56 & 0.80 & 0.17 & 0.48 \\
DMDR & 4 & 4 & 0.64 & \textbf{0.99} & 0.84 & 0.54 & \textbf{0.81} & 0.25 & 0.44 \\
\rowcolor[RGB]{240,230,245}
GDMD (Ours) & 4 & 4 & \textbf{0.71} & \textbf{0.99} &  \textbf{0.86} & \textbf{0.71} & \textbf{0.81} & \textbf{0.30} & \textbf{0.62} \\

\arrayrulecolor{black}\bottomrule
\end{tabular}}
\vspace{-0.2em}
\label{tab:geneval}
\end{table*}

%% file: fig/user-study.tex
\begin{figure}[!b]
  \centering
    \includegraphics[width=\linewidth]{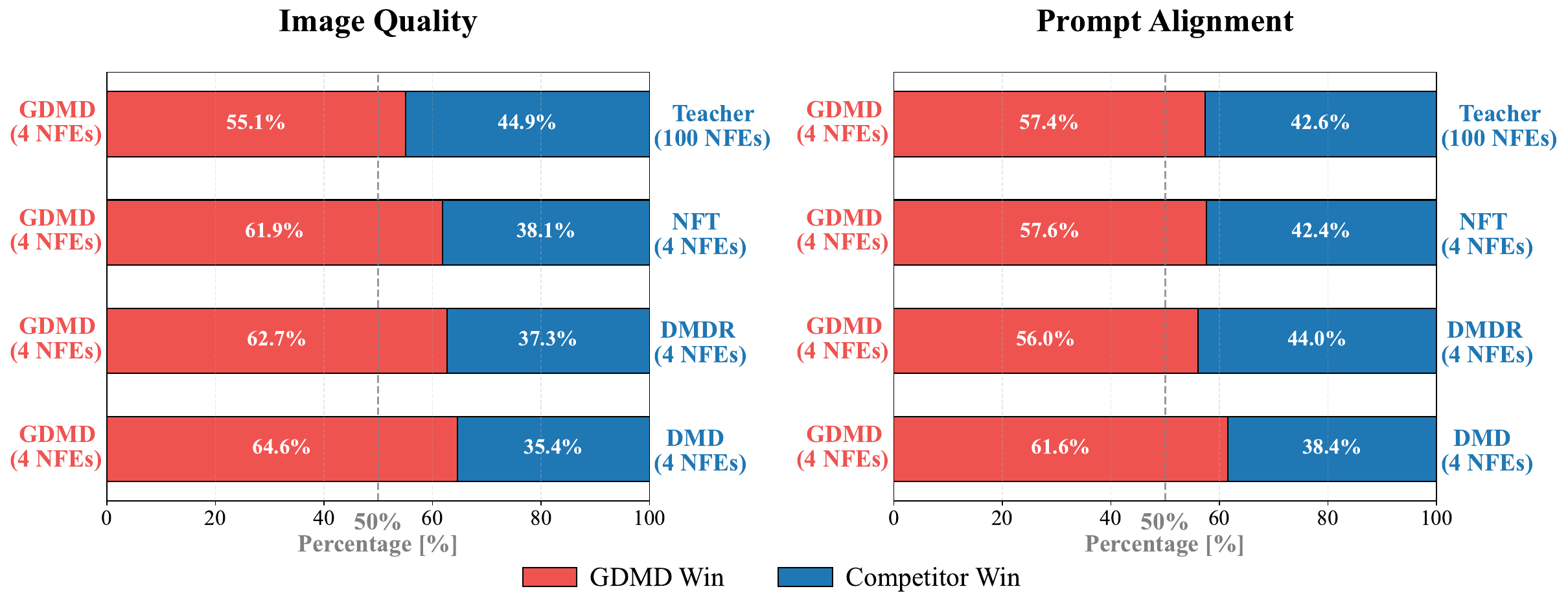}
   \caption{
   \textbf{Human preference study.}
   We compare the performance of GDMD (4-step) against established baselines. 
   Our GDMD model outperforms all other models, including its teacher, DMD, DMDR and DMD + NFT 
   in human preference, for both image quality
    and prompt alignment.
    }
   \label{fig:user-study}
\end{figure}

%% file: fig/ab_vis.tex
\begin{figure}[!t]
  \centering
    \includegraphics[width=\linewidth]{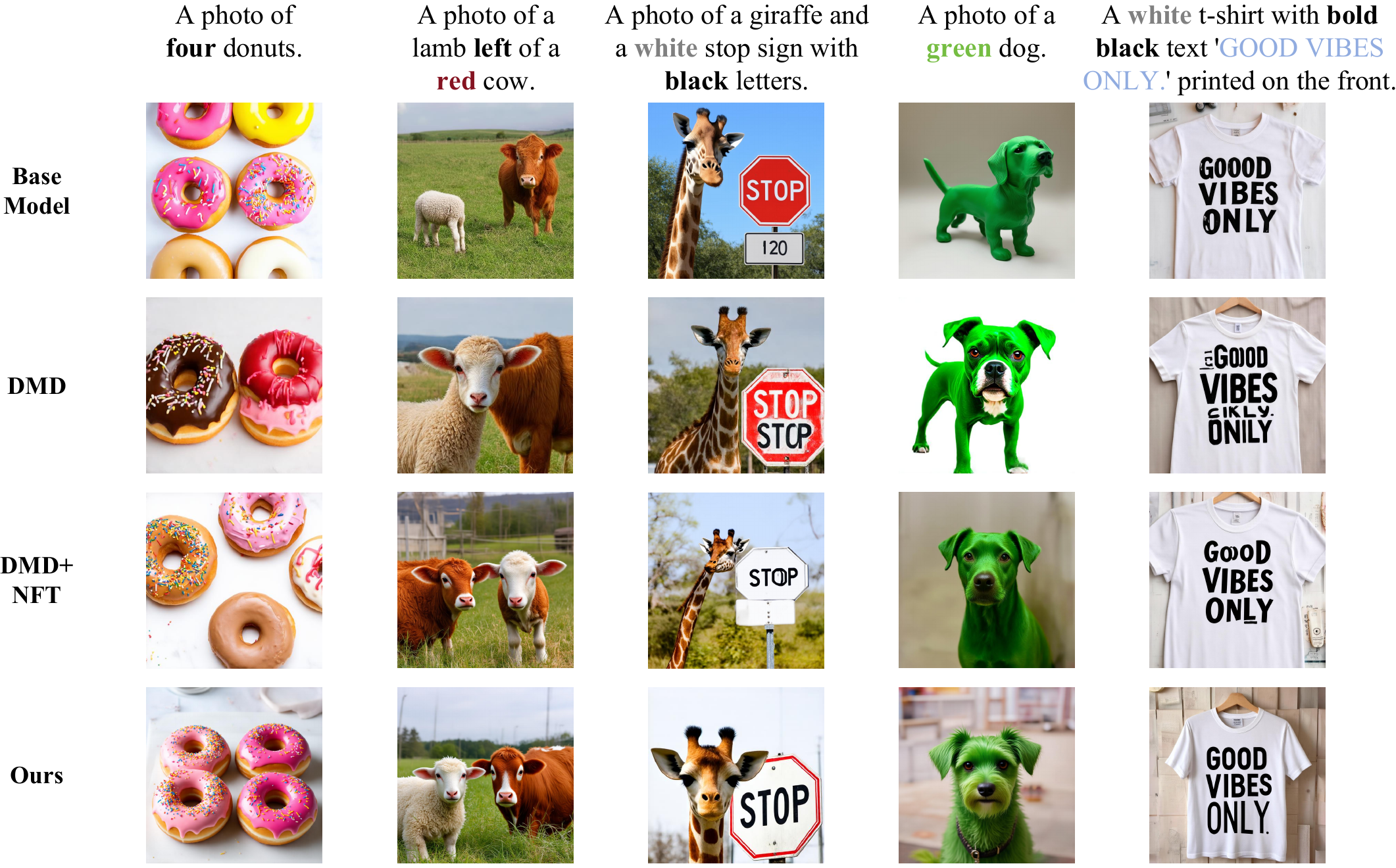}
   \caption{Visual comparison of different tasks (Counting, Position, Attribute Binding, Colors and Texts) across different methods: Teacher, DMD and DMD + NFT. 
   GDMD outperforms others in both quality and prompt alignment, showing strong performance.
   }
   \vspace{-0.3em}
   \label{fig:ab_vis}
\end{figure}

%% file: tables/ab_rl.tex
\begin{table}[!t]
\centering
\captionof{table}{\textbf{Ablation study on different RL strategies.} 
All RL methods are trained solely on Clip Score and HP Score. The best result is highlighted in \textbf{bold}.}
\resizebox{1\linewidth}{!}{
\begin{tabular}{l c c c c c}
\toprule
\multirow{2}{*}{\textbf{Method}} & \multicolumn{2}{c}{\textbf{Trained Rewards}} & \multicolumn{3}{c}{\textbf{Unseen Rewards}} \\
\cmidrule(lr){2-3} \cmidrule(lr){4-6}
& ~CLIP Score$\uparrow$ & ~HP Score$\uparrow$ & ~Aesthetic Score$\uparrow$ &~Pick Score$\uparrow$ & ~ImageReward$\uparrow$\\
\midrule
Base-Model (w/ CFG) & 0.2936 & 0.2810 & 5.5711 & 22.3236 & 1.0759 \\
\arrayrulecolor{black!40}\midrule
w/ only Distill. & 0.2861 & 0.2891 & 5.5598 & 21.6216 & 0.9704 \\
w/ Distill. + RL (ReFL) & {0.2901} & {0.2931} & {5.5123} & {21.6528} & {1.0120}  \\
w/ Distill. + RL (GRPO) & 0.2929 & 0.2963 & 5.5854 & 22.3574 & 1.1329\\
w/ Distill. + RL (NFT) & \textbf{0.2931} & 0.2894 & 5.7446 & 22.3782 & 1.1648 \\
\rowcolor[RGB]{240,230,245}
GDMD (Ours) & {0.2930} & \textbf{0.3076} & \textbf{5.8728} & \textbf{22.4614} & \textbf{1.2702} \\
\arrayrulecolor{black}\bottomrule
\end{tabular}}
\vspace{-0.5em}
\label{tab:ab_rl}
\end{table}

%% file: tables/ap/ab_trick.tex
\begin{table}[t]
\centering
\captionof{table}{\textbf{Ablation study of different gradient collection strategies.}}
\resizebox{1\linewidth}{!}{
\begin{tabular}{l c c c c c}
\toprule
\multirow{2}{*}{\textbf{Method}} & \multicolumn{2}{c}{\textbf{Trained Rewards}} & \multicolumn{3}{c}{\textbf{Unseen Rewards}} \\
\cmidrule(lr){2-3} \cmidrule(lr){4-6}
& ~CLIP Score$\uparrow$ & ~HP Score$\uparrow$ & ~Aesthetic Score$\uparrow$ &~Pick Score$\uparrow$ & ~ImageReward$\uparrow$\\
\midrule
Base-Model (w/ CFG) & 0.2936 & 0.2810 & 5.5711 & 22.3236 & 1.0759 \\
\arrayrulecolor{black!40}\midrule
DMD & 0.2861 & 0.2891 & 5.5598 & 21.6216 & 0.9704 \\
GDMD (Ours) / diff. $t$ & 0.2916 & 0.3005 & 5.8175 & 22.2846 & 1.2363 \\
GDMD (Ours) / diff. $\mu_{fake}^\phi$ & 0.2879 & 0.2983 & 5.8126 & \textbf{22.4640} & 1.2651 \\
\rowcolor[RGB]{240,230,245}
GDMD (Ours) / ALL. & \textbf{0.2930} & \textbf{0.3076} & \textbf{5.8728} & 22.4614 & \textbf{1.2702} \\

\arrayrulecolor{black}\bottomrule
\end{tabular}}
\label{tab:ab_trick}
\end{table}

%% file: section/5_conclusion.tex
\section{Conclusion}
\label{sec:conclu}
In this paper, we present {GDMD}, a novel framework that redefines the synergy between DMD and RL by shifting the reward paradigm from raw sample pixels to \textbf{gradient-based scoring}. 
By leveraging DMD optimization gradients as the primary guidance signal, our approach effectively overcomes the optimization instabilities and noisy feedback loops inherent in traditional sample-based scoring,
particularly during the early stages of few-step distillation. 
Through the integration of {Distillation-aware Gradient Collection}, {Implicit Gradient Scoring} and {Negative-aware Preference Optimization}, {GDMD} eliminates the need for a cold-start phase and expensive SDE sampling, ultimately establishing a new state-of-the-art in generative efficiency and human preference alignment. 
Extensive benchmarks show that our method bridges the distillation quality gap, outperforms existing baseline models, and offers a robust, scalable solution for high-quality image synthesis.

\paragraph{Limitation and Furture Work.} 
Although extensive experiments have demonstrated that our method achieves high quality generation, the method itself is limited by the quality of the reward model. 
When score evaluation fails, distillation optimization drifts to low-quality directions, thereby limiting performance.
Since GDMD jointly trains DMD with reinforcement learning, a promising direction is to apply GDMD to long video generation, where its joint training scheme could mitigate drift and degradation and further improve generation quality over long temporal horizons.


%% file: section/6_acknowledgement.tex
\clearpage
\section*{Acknowledgments}
This work was supported by the National Natural Science Foundation of China under Grant U2336212, and by the "Pioneer" and "Leading Goose" R\&D Program of Zhejiang under Grant 2024C01161.

%% file: section/X_appendix.tex
\appendix
\counterwithin{figure}{section}
\counterwithin{table}{section}
\counterwithin{equation}{section}

\section*{Supplementary Material}
\input{section/appendix/A_method}
\input{section/appendix/B_setup}

\input{section/appendix/C_results}

\input{section/appendix/D_ablations}
\input{section/appendix/E_qualitative}
\input{section/appendix/F_impact}

%% file: section/appendix/A_method.tex
\section{Methodological Details}
\label{sec:ap:method}
\subsection{Further Analysis of GDMD}
\label{sec:proof}
GDMD utilizes the KL divergence signal between the real and fake distributions as the basis for evaluation and optimization.
Following DMD, We minimize the Kullback–Leibler (KL) divergence between the distributions of real and fake images, denoted as $p_\text{real}$ and $p_\text{fake}$:
\begin{align}
D_{KL}(p_{\text{fake}} \parallel p_{\text{real}}) &= \mathbb{E}_{x \sim p_{\text{fake}}} \left( \log \left( \frac{p_{\text{fake}}(x)}{p_{\text{real}}(x)} \right) \right) \nonumber \\
&= \mathbb{E}_{z \sim \mathcal{N}(0; I)} \left[ -\left( \log p_{\text{real}}(x) - \log p_{\text{fake}}(x) \right) \right], 
\label{ap:eq:kl}
\end{align}
where $x = G_{\theta}(z)$.
Let $s_{\text{real}}(x) = \nabla_x \log p_{\text{real}}(x), \quad s_{\text{fake}}(x) = \nabla_x \log p_{\text{fake}}(x)$ denote the scores of the respective distributions. We then study the expression in \cref{ap:eq:kl} for the gradient with respect to the parameter $\theta$:
\begin{align}
\nabla_{\theta} D_{KL} 
&\simeq \mathbb{E}_{z, t, x, x_t} \left[ w_t \left( s_{\text{fake}}(x_t, t) - s_{\text{real}}(x_t, t) \right) \frac{\partial x_t}{\partial \theta} \right] \nonumber \\
&= \mathbb{E}_{z, t, x, x_t} \left[ w_t \left( s_{\text{fake}}(x_t, t) - s_{\text{real}}(x_t, t) \right) \frac{\partial x_t}{\partial G_{\theta}(z)} \frac{\partial G_{\theta}(z)}{\partial \theta} \right]. \nonumber \\
&= \mathbb{E}_{z, t, x, x_t} \left[ w_t \left( s_{\text{fake}}(x_t, t) - s_{\text{real}}(x_t, t) \right) \frac{\partial x_t}{\partial x} \frac{\partial G_\theta(z)}{\partial \theta} \right] \nonumber \\
&= \mathbb{E}_{z, t, x, x_t} \left[ w_t \, \alpha_t \left( s_{\text{fake}}(x_t, t) - s_{\text{real}}(x_t, t) \right) \frac{dG_\theta(z)}{d\theta} \right] \nonumber \\
& =  w_t \, \alpha_t \, \mathbb{E}_{z, t, x, x_t} \left[ \left( s_{\text{fake}}(x_t, t) - s_{\text{real}}(x_t, t) \right) \frac{dG_\theta(z)}{d\theta} \right] \nonumber \\
& = \frac{1}{2} w_t \, \alpha_t \nabla_\theta \mathbb{E}_{z, t, x, x_t} \left|\left|G_\theta(z) - (G_{\theta^{-}}(z) -  \left( s_{\text{fake}}(x_t, t) - s_{\text{real}}(x_t, t) \right)  \right|\right|^2_2,
\label{ap:eq:der}
\end{align}
where $x_t$ is the noisy version of $x$ at time $t$.
Through the DMD equivalence transformation, \cref{ap:eq:der} expresses the gradient optimization in terms of an MSE loss.
Evaluating the implicit representation of target $x_\text{tar} := (G_{\theta^{-}}(z) -  \left( s_{\text{fake}}(x_t, t) - s_{\text{real}}(x_t, t) \right)$ indicates the quality of the update.

\subsection{Further Analysis of DMD with RL}
\input{fig/ap/scatter}
To gain deeper insight into the behavior of DMD, we visualize and compare the feature spaces of the teacher model and the DMD student, as shown in \cref{fig:scatter}.
The teacher model shows a similar broader distribution in color (``Red'' or `Blue''), with clearer separation between object (``Car'', ``Cat'' or ``Dog'') clusters.
However, DMD's distributions can suffer from confusion, which manifests as outliers (marked with $\star$) in the categories ``Cat'' and ``Dog''.
These outliers are often accompanied by increased oversaturation, distortion, or blurring, as shown in \cref{fig:scatter} (b).
While previous GAN-based methods resolve this distribution confusion through adversarial training with real data, 
reinforcement learning (RL) presents a distinct alternative by specifically shaping the distribution of outliers through preference-based control, as shown in \cref{fig:scatter} (c).

Building upon on the analysis of prior research, integrating Reinforcement Learning (RL) into the Distribution Matching Distillation (DMD) pipeline effectively overcomes the performance ceiling and mode-collapse issues inherent in vanilla distillation.
While standard DMD restricts the student to a mere imitation of the teacher’s distribution, the joint training framework introduces reward-driven supervision (targeting aesthetic quality and semantic alignment), enabling the student to transcend the teacher’s capabilities without requiring external image data.
As explained by DMDR, by actively steering the student’s manifold toward high-reward regions, the RL objective facilitates an escape from the teacher's suboptimal modes and mitigates ``zero-forcing'' artifacts, thereby ensuring superior coverage of preferred image modes, as shown in \cref{fig:ap:mode} left.
Simultaneously, DMD acts as a potent regularizer for the RL process, effectively suppressing reward hacking and rapid overfitting.
This dual-objective ensures that while the student explores high-reward spaces, it remains grounded within the structural priors of the teacher, stabilizing the optimization trajectory and reducing error accumulation.

\subsection{Further Analysis of Implicit Targets}
\label{sec:ap:target}
To assess whether implicit targets provide reliable preference signals,
we collected 10,000 paired student outputs $x_0$ and implicit targets $x_{\mathrm{tar}}$ using object-counting-style prompts and different random seeds during the first 1,000 training steps. 
Following DMD2~\cite{yin2024improved}, we sampled $x_t$ through backward simulation and obtained $x_0$ by predicting the clean sample at timestep zero with
the student model. 
The corresponding implicit targets $x_{\mathrm{tar}}$ were constructed using DaGC.

\Cref{tab:rebuttal:target} compares the decoded samples in terms of image quality, text--image alignment, and downstream detection performance. 
Implicit targets outperform student outputs across all six metrics. 
In particular, object-detection and counting-detection rates increase from 50.71\% to 58.15\% and from 31.35\% to 36.12\%, respectively. 
Moreover, 74.3\% of the implicit targets receive a higher CLIP score than their paired student outputs and are therefore accepted as positive samples under the CLIP-based criterion.
\Cref{fig:visual_comparison} presents examples of accepted and rejected samples during early training.

\input{tables/rebuttal/target}


To further examine the effect of the scoring strategy, the matched-policy comparison in \cref{tab:ab_rl} contrasts DiffusionNFT with $x_0$-based scoring against GDMD with implicit-target scoring.
The sample-scoring variant uses an equally weighted combination of CLIP and HPS rewards. 
Target scoring improves ImageReward from 1.1648 to 1.2702 and HPS from 0.2894 to 0.3076, while yielding a slightly lower CLIP score (0.2930 versus 0.2931). 
Under the matched preference-learning policy, these results provide additional evidence for the benefit of scoring implicit targets.
\input{fig/ap/ap_select_show}

%% file: fig/ap/scatter.tex
\begin{figure}[!t]
  \centering
    \includegraphics[width=\linewidth]{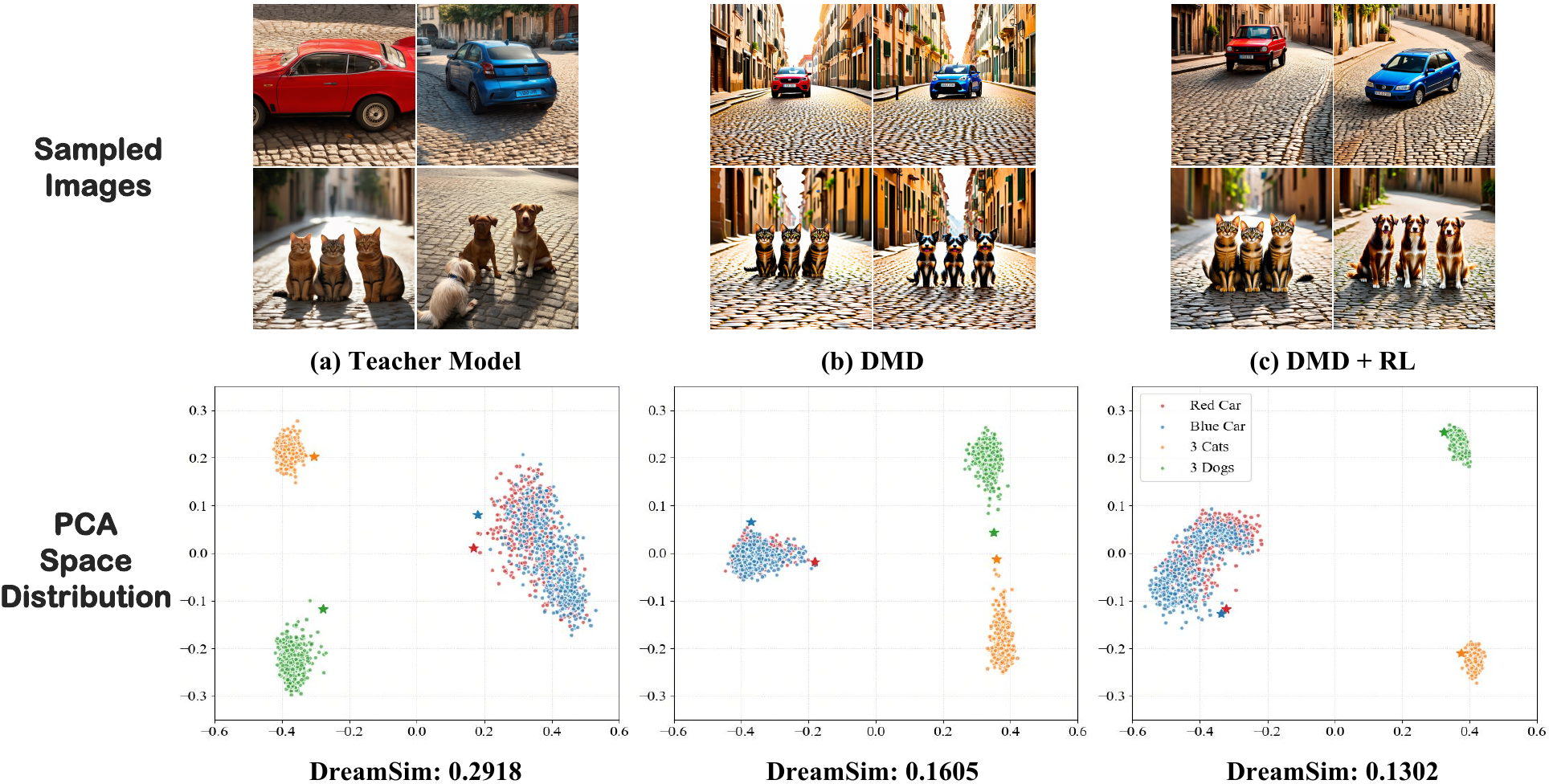}
    \caption{\textbf{Distribution analysis and visualization.} 
    We compare the feature of the Teacher, DMD, and DMD+RL. 
    Representative outlier images (marked with $\star$) are shown in the top row. 
    The Teacher model exhibits high diversity but includes distorted samples. 
    DMD imposes constraints but introduces over-saturation and artifacts. 
    In contrast, DMD+RL refines the optimization using reward feedback, pulling outliers toward the cluster center, yielding a more compact cluster with lower DreamSim scores (Averaged over 4 clusters; lower is tighter).
    }
   \label{fig:scatter}
\end{figure}

%% file: tables/rebuttal/target.tex
\begin{table}[!t]
\centering
\caption{\textbf{Early-training comparison of student samples and implicit targets.}
Evaluation uses 10,000 collected samples. Higher is better for all reported measures.}
\label{tab:rebuttal:target}
\resizebox{\linewidth}{!}{%
\begin{tabular}{l c c c c c c}
\toprule
\textbf{Target}
& CLIP Score$\uparrow$
& HP Score$\uparrow$
& Aesthetic Score$\uparrow$
& Pick Score$\uparrow$
& Object.$\uparrow$
& Count.$\uparrow$ \\
\midrule
$x_0$
& 0.2593 & 0.1959 & 4.7347 & 0.7852 & 50.71\% & 31.35\% \\
\rowcolor[RGB]{240,230,245}
$x_{\mathrm{tar}}$
& \textbf{0.2630}
& \textbf{0.2018}
& \textbf{4.7712}
& \textbf{0.7964}
& \textbf{58.15\%}
& \textbf{36.12\%} \\
\arrayrulecolor{black}\bottomrule
\end{tabular}%
}
\par\smallskip
\begin{minipage}{\linewidth}
\small
Object. and Count. denote the reported object-detection and
counting-detection rates. 
\end{minipage}
\end{table}

%% file: fig/ap/ap_select_show.tex
\begin{figure}[!t]
  \centering

  \includegraphics[width=0.15\linewidth]{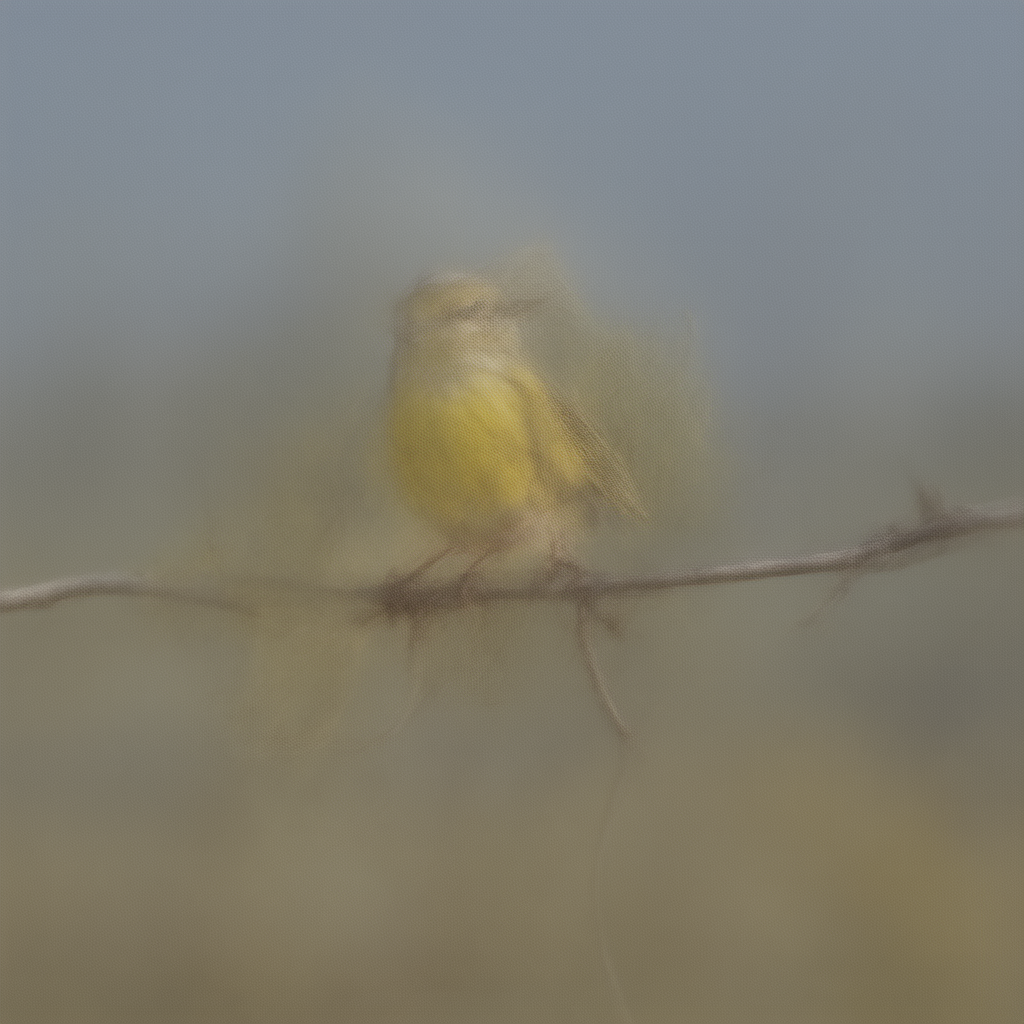}\hfill
  \includegraphics[width=0.15\linewidth]{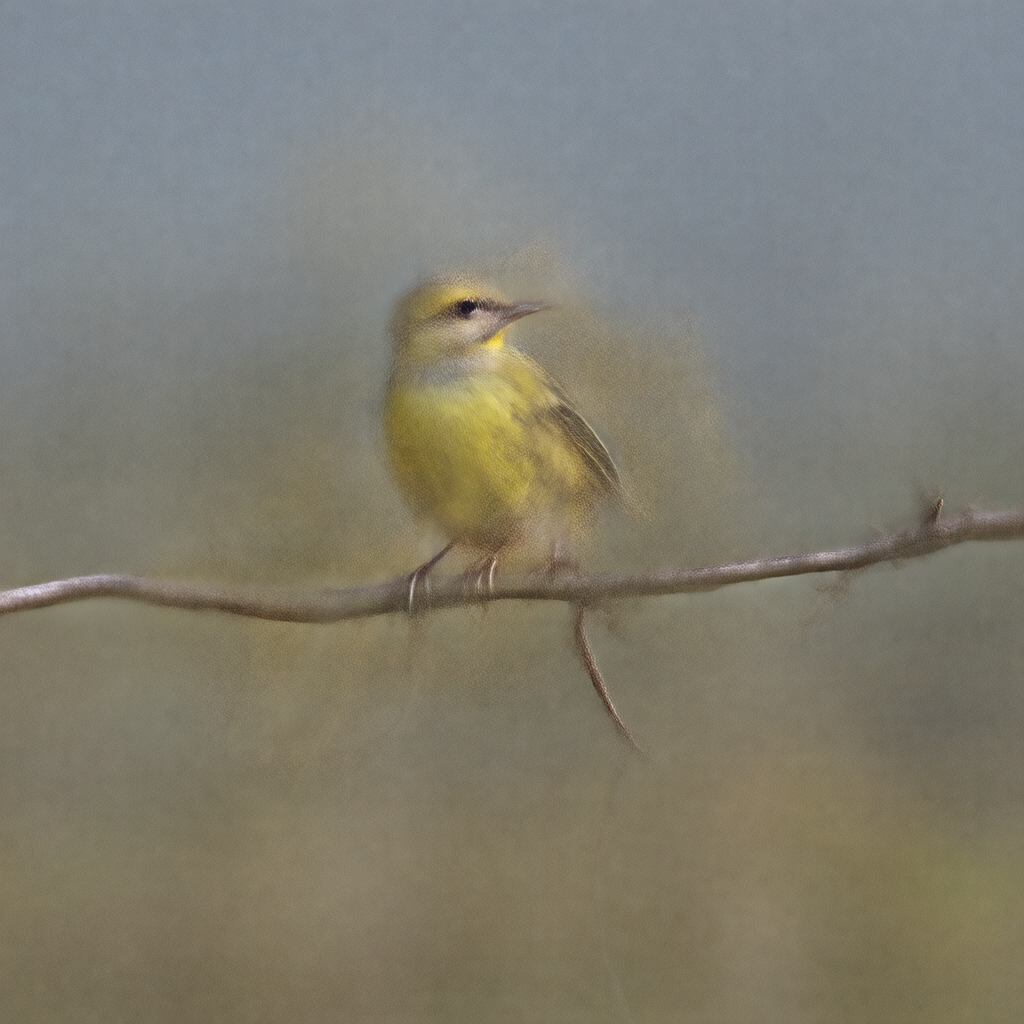}\hfill
  \includegraphics[width=0.15\linewidth]{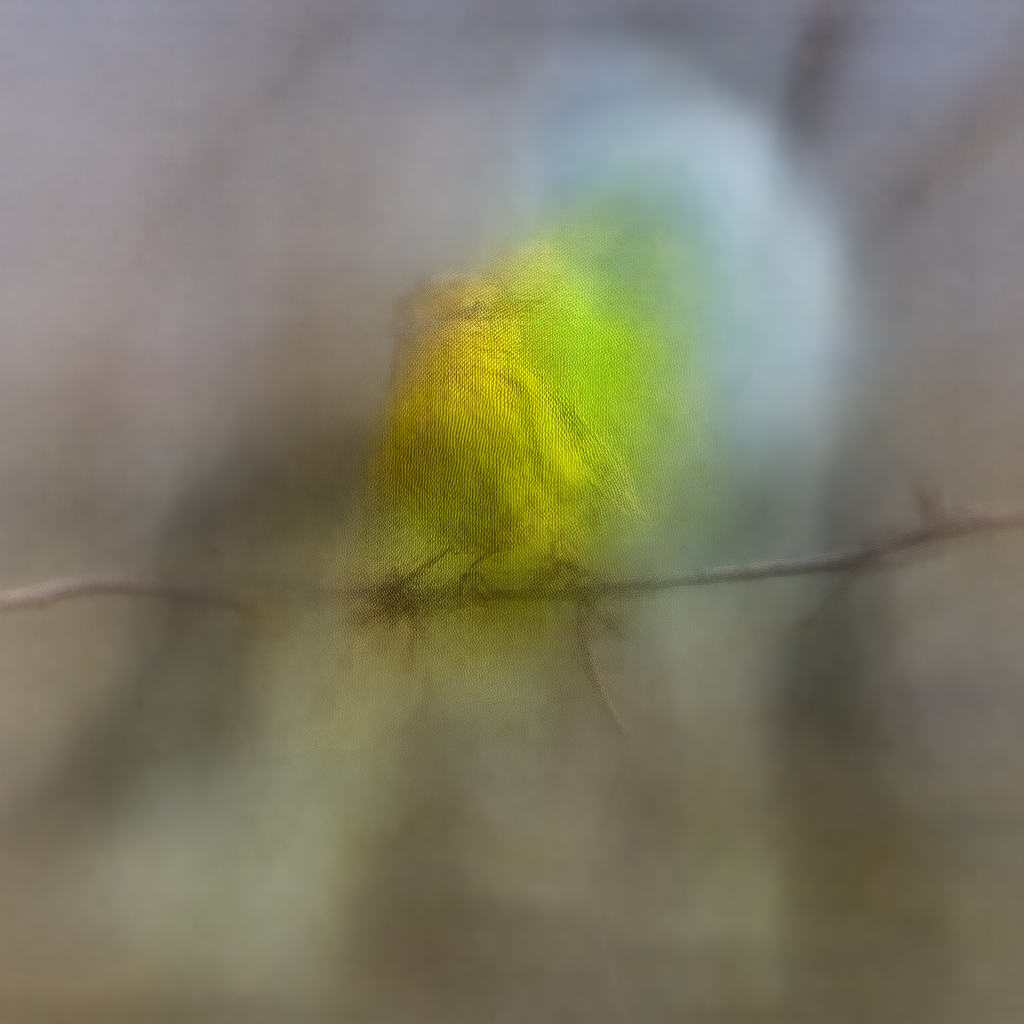}%
  \hspace{0.04\linewidth}%
  \includegraphics[width=0.15\linewidth]{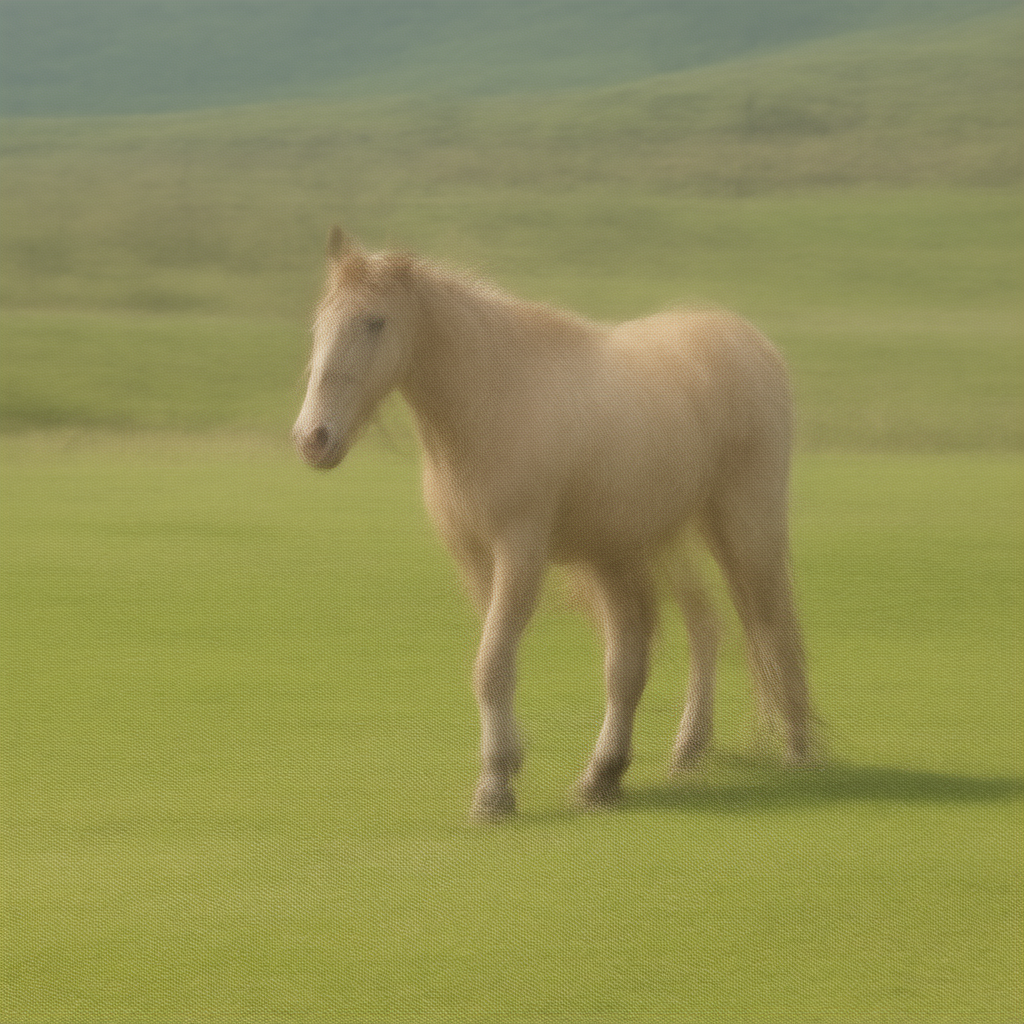}\hfill
  \includegraphics[width=0.15\linewidth]{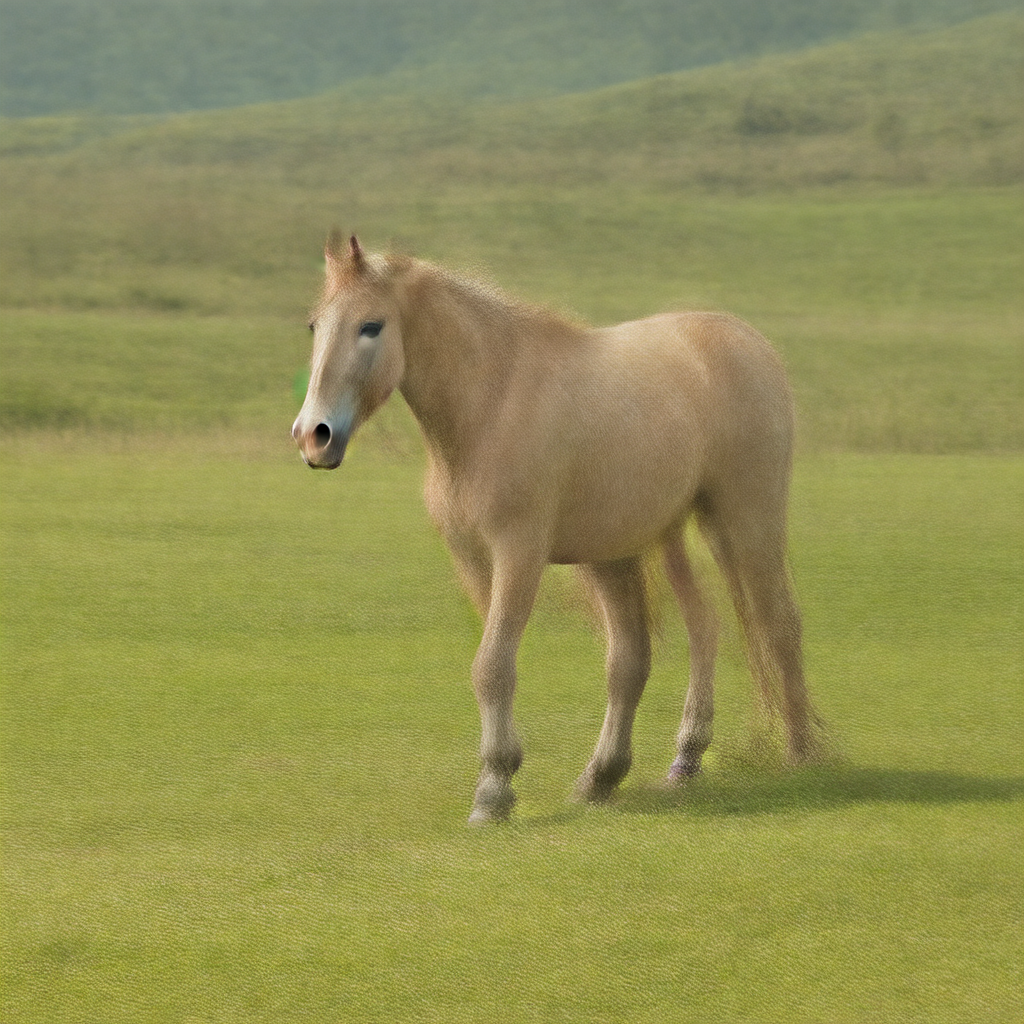}\hfill
  \includegraphics[width=0.15\linewidth]{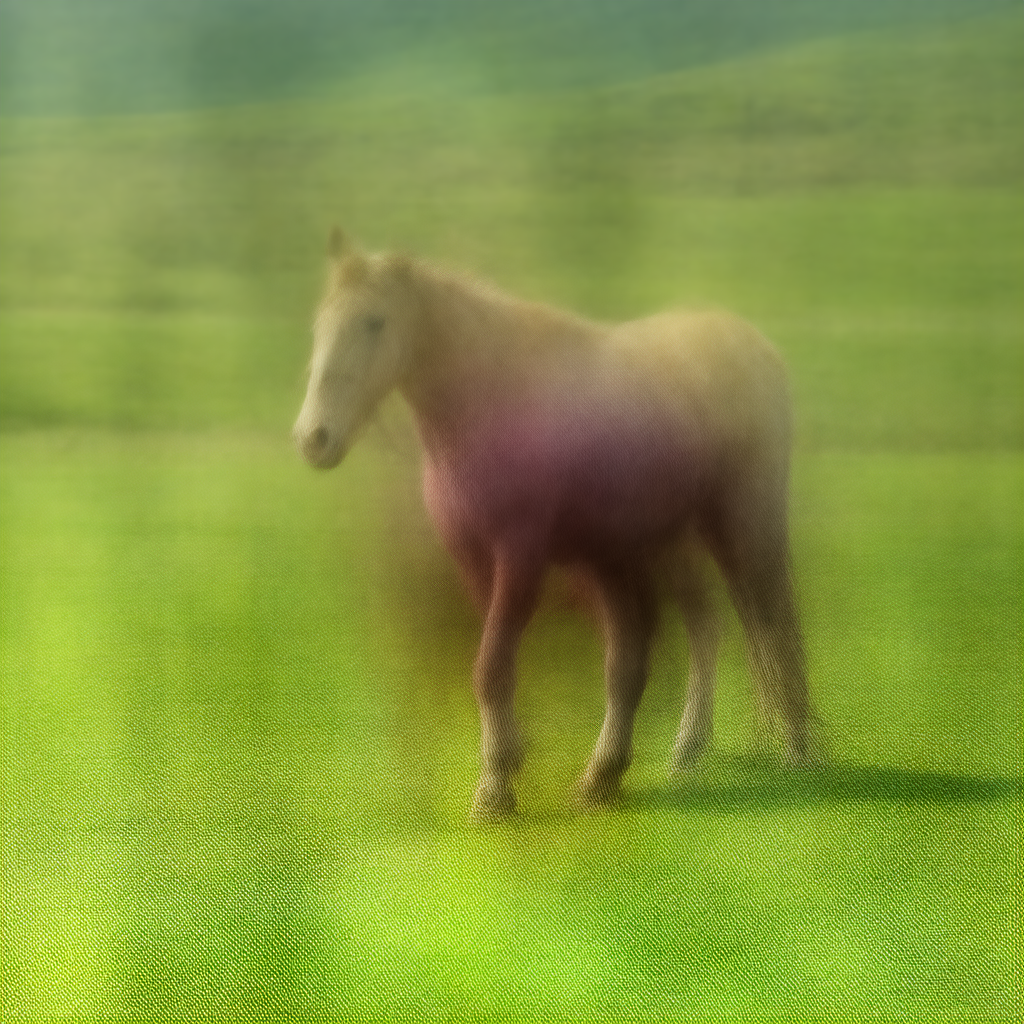}%
  \par\vspace{4pt}

  \includegraphics[width=0.15\linewidth]{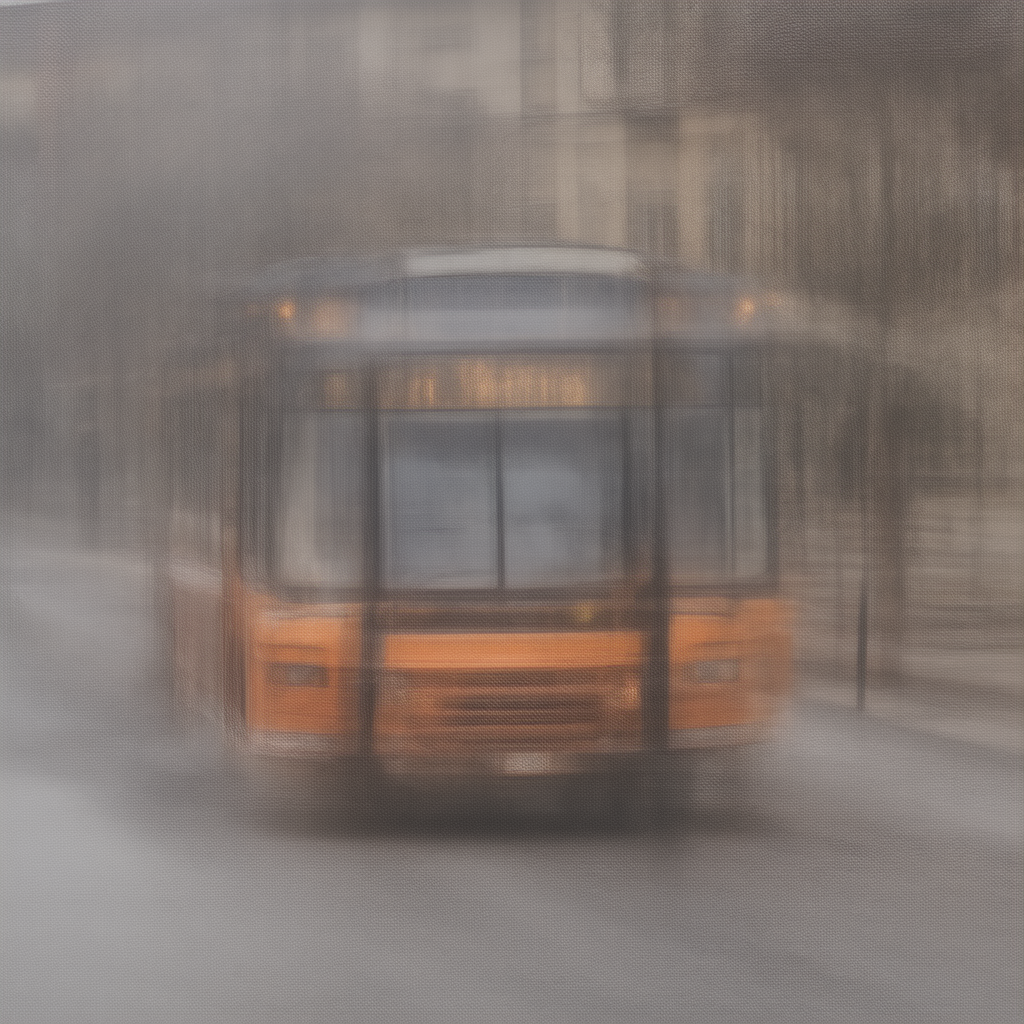}\hfill
  \includegraphics[width=0.15\linewidth]{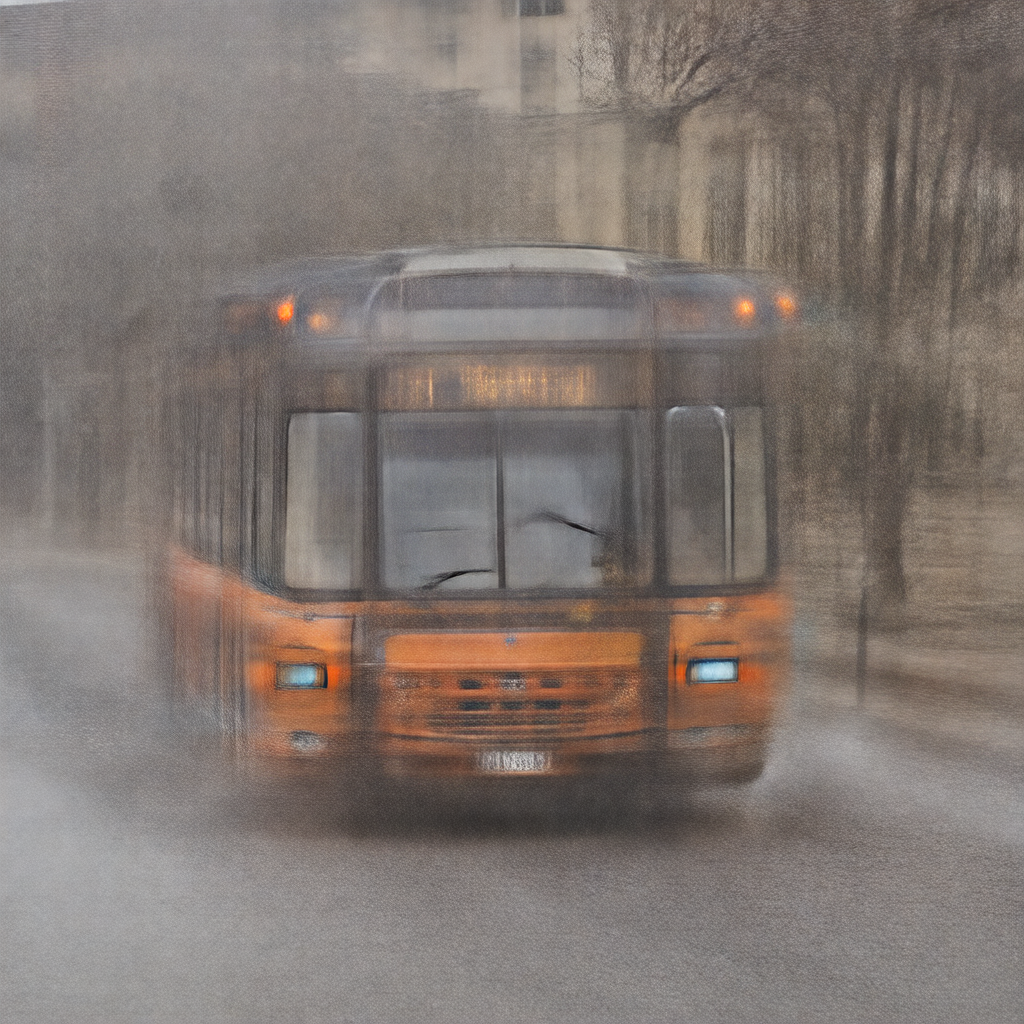}\hfill
  \includegraphics[width=0.15\linewidth]{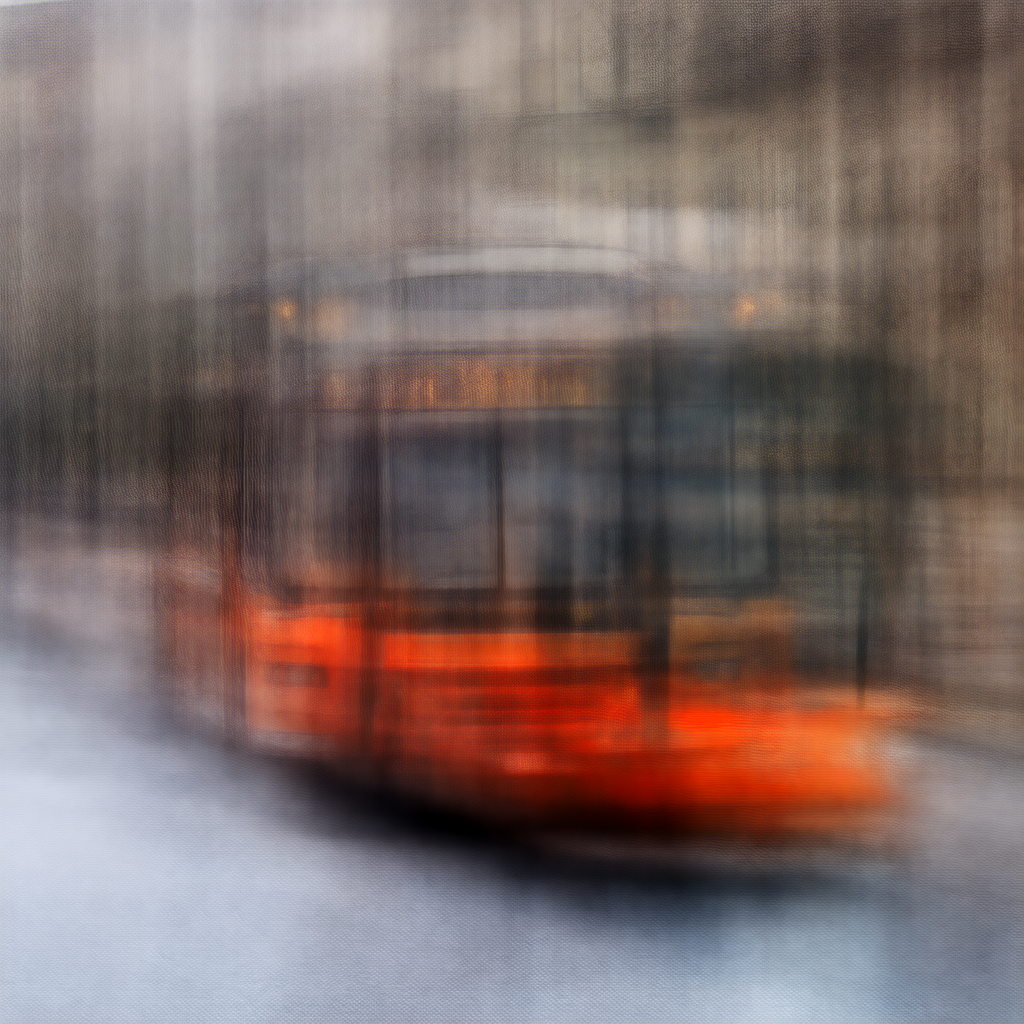}%
  \hspace{0.04\linewidth}%
  \includegraphics[width=0.15\linewidth]{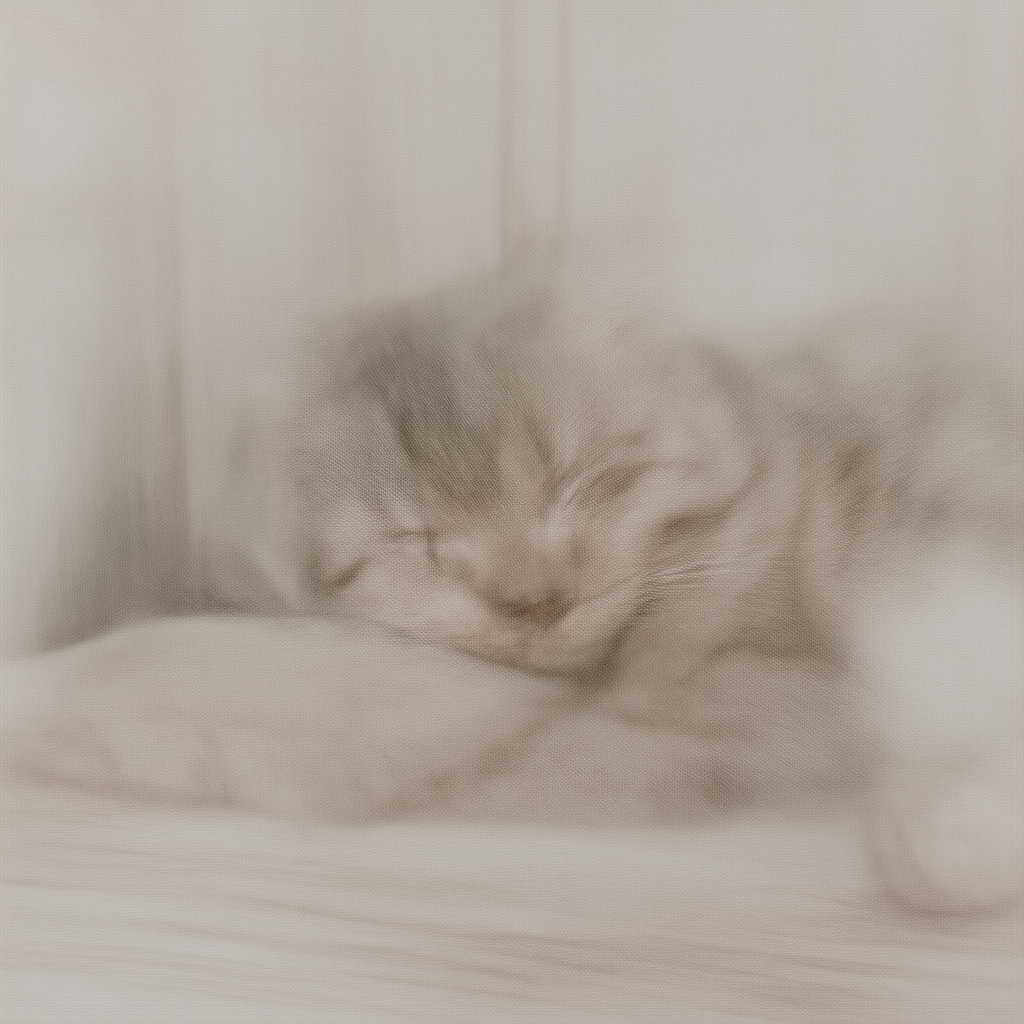}\hfill
  \includegraphics[width=0.15\linewidth]{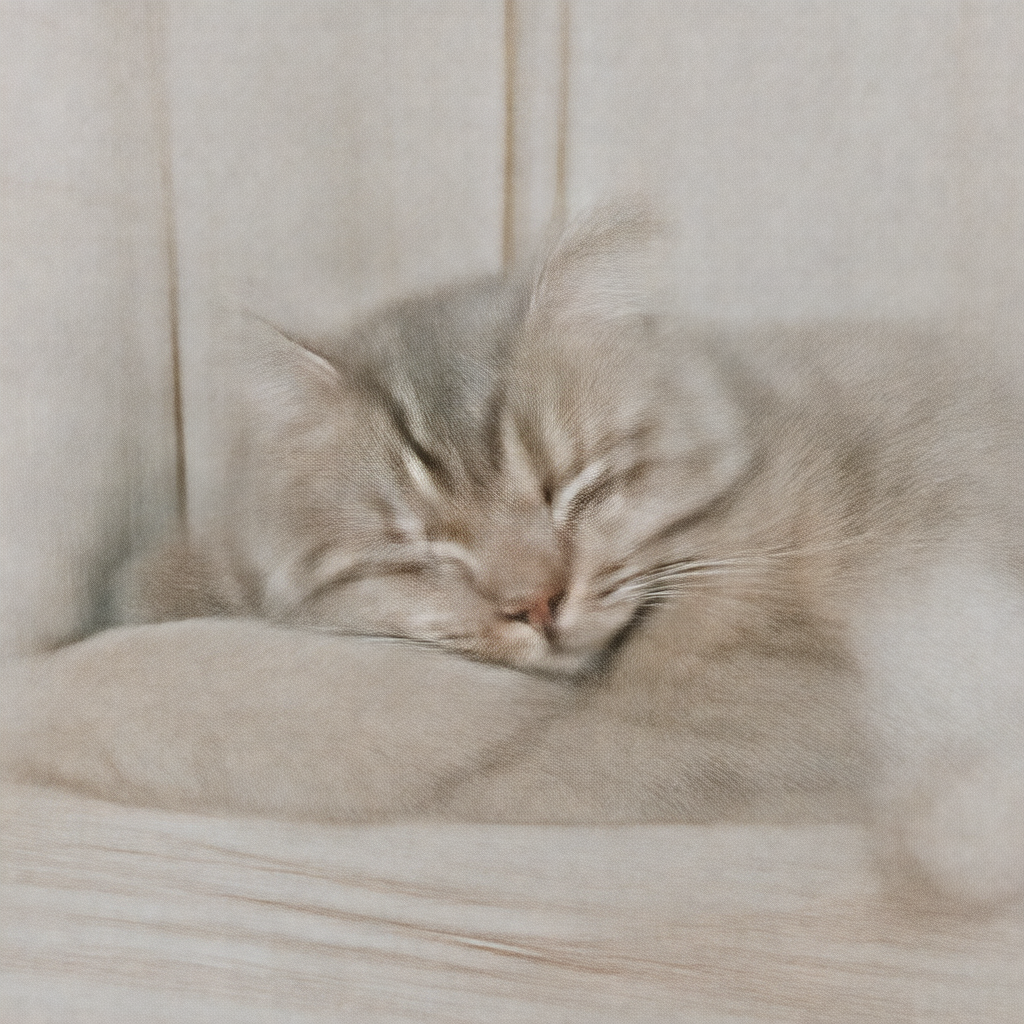}\hfill
  \includegraphics[width=0.15\linewidth]{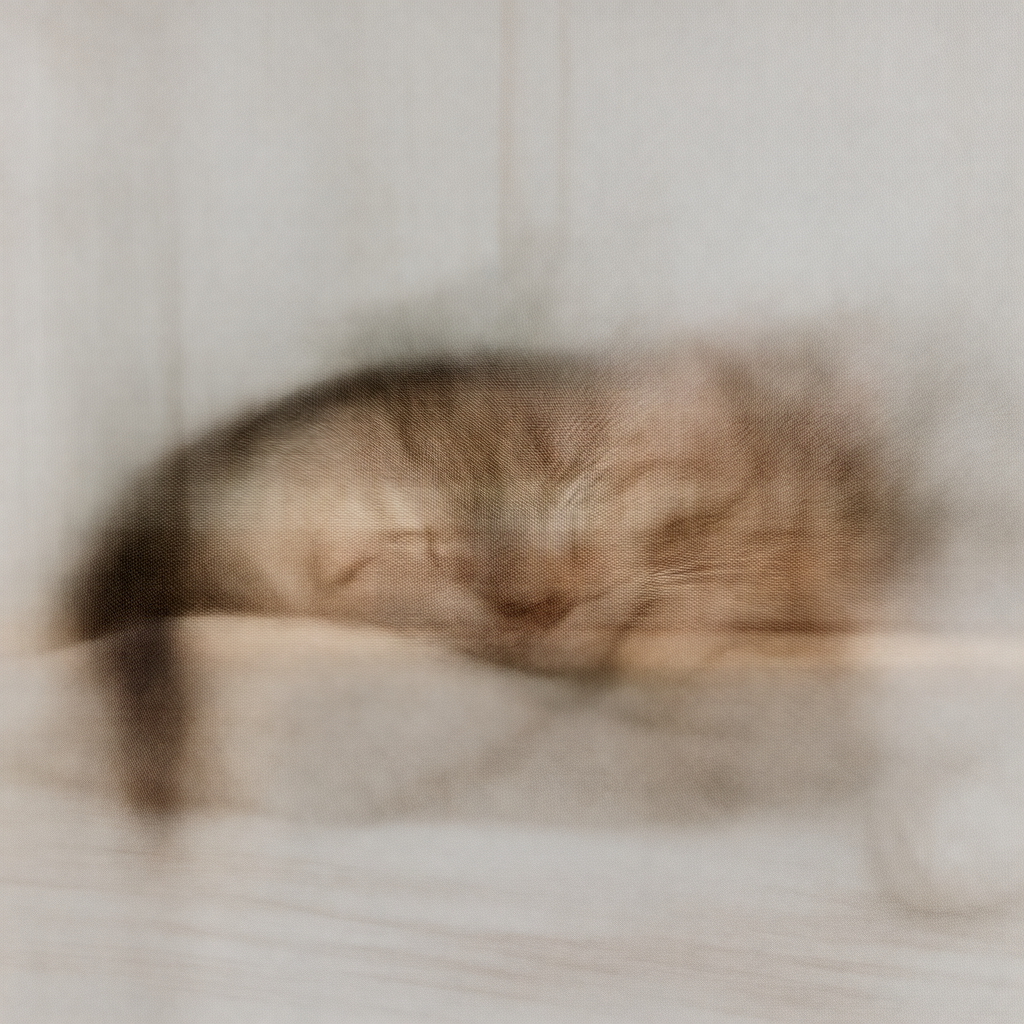}%
  \par\vspace{4pt}



  \makebox[0.15\linewidth]{\small $x_0$}\hfill
  \makebox[0.15\linewidth]{\small\textcolor{green!50!black}{Accept}}\hfill
  \makebox[0.15\linewidth]{\small\textcolor{red}{Reject}}%
  \hspace{0.04\linewidth}%
  \makebox[0.15\linewidth]{\small $x_0$}\hfill
  \makebox[0.15\linewidth]{\small\textcolor{green!50!black}{Accept}}\hfill
  \makebox[0.15\linewidth]{\small\textcolor{red}{Reject}}
  \par
  \caption{\textbf{Visual comparison of accepted and rejected outputs.}
    Each group shows the generator output's $x_0$, an accepted output,
    and a rejected output, from left to right.}
  \label{fig:visual_comparison}
\end{figure}

%% file: section/appendix/B_setup.tex
\section{Implementation Details}
\label{sec:ap:imple}
\subsection{DMD Configuration}
Following DMD, we employ the same adaptive weight calculation to facilitate a fair comparison:
\begin{equation}
w_t =  \frac{CS}{\| \mu_{\text{real}}(x_t, t) - x_0 \|_1},
\end{equation}
where $S$ is the number of spatial locations and $C$ is the number of channels.
For the Renoise scheduler, let the total number of timesteps be 
$T$.
We similarly sample $t \sim \mathcal{U}(T_\text{min}, T_\text{max})$, where $T_\text{min} = 0.02 T, T_\text{max} = 0.98 T$, as defined in DMD.
In our approach, classifier-free guidance (CFG) is applied solely to the real score function for distillation, as in DMD, while the fake score remains the model's standard prediction.
For image-free data distillation, we construct noisy data by simulating the backward process.
We employ LoRA to train both the generator and the fake score function. 
To alleviate GPU memory constraints, we use the base model as the real score function, while its LoRA-adapted variant serves as the fake score function.
The Lora rank has been set to 64.

\subsection{Preference Optimization}
The RL policy for group-style requires score normalization, typically using the mean and variance of a dataset, such as that employed in DiffusionNFT:
\begin{equation}
r(x_0, c) := \frac{1}{2} + \frac{1}{2} \operatorname{clamp} \left[ \frac{r^{\text{raw}}(x_0, c) - \mathbb{E}_{\pi_{\text{old}}(\cdot|c)} r^{\text{raw}}(x_0, c)}{Z_c}, -1, 1 \right].
\end{equation}
In practice, we have found that the score of $x_0$ sampled under the current optimization state can serve as a benchmark for evaluating the quality of gradients.
Therefore, we adopt Eq. 5 as the final normalization method.
For RL-based optimization, we employ DiffusionNFT, as its positive-negative sample concept naturally aligns with our requirement for both positive and negative gradient signals (See \cref{fig:ap:mode} right).
For the adaptive weights employed, we adopt a similar regularization approach:
\begin{equation}
w(t) = \frac{1}{\operatorname{sg}\left( \operatorname{mean}\left( \operatorname{abs}\left( x_\theta(x_t, c, t) - x_0 \right) \right) \right)}
\end{equation}
For old strategy construction, we employ two approaches. 
The first utilizes the Soft Update method:
\begin{equation}
\theta^{\text{old}} \leftarrow \eta_i \, \theta^{\text{old}} + (1 - \eta_i) \, \theta
\end{equation}
Another approach is to replace the old strategy with an estimate using the fake score, since the fake score is a lagging update of the generator.

\subsection{Training Configuration}
We train GDMD for approximately 10 hours on 8 NVIDIA A100 80GB GPUs, using a batch size of 2 per GPU (total batch size 16). We adopt the AdamW optimizer with a learning rate of 3e-5, $(\beta_1, \beta2)=(0.9,0.999)$, and a weight decay of 1e-4. 
No gradient accumulation is used, and we synchronize gradients across all GPUs using DistributedDataParallel. 
The training comprises approximately 10k steps.
We set the DMD update ratio to 5, the DiffiusionNFT hyperparameter $\beta$ to 1, and the loss hyperparameters $\lambda$ and $\gamma$ to 1 and 0.1, respectively.

\subsection{Training Algorithm}
\label{sec:ap:algo}
\input{algo/algo}
\Cref{alg:distillation} shows our entire training process.

\subsection{Human Evaluation Setting}
We conducted a user study to assess the generated images from two perspectives: visual quality and text‑image alignment. Each participant was presented with 20 distinct text prompts. For each prompt, five images generated by different models (labeled A, B, C, D, and E) were shown side by side. The order of the images was randomized per prompt to avoid positional bias.
For each set of images, participants answered two multiple‑choice questions (one or more options could be selected):
\begin{itemize}
    \item Image Quality – Which image(s) exhibit better visual quality?
(e.g., clearer, better composition, more natural lighting, richer details)
    \item Text Match – Which image(s) best match the given text description?
\end{itemize}
Participants were instructed to base their judgments on first impressions and intuition, without overthinking. The study was conducted online, and each session took approximately 10–15 minutes to complete.
The human evaluation involved 137 participants, each of whom was paid \$0.10 per evaluation.

The user study described in \cref{sec:exp_human} involved only image evaluation tasks with no sensitive or personally identifiable information collected. Participants were informed that the task was voluntary and that they could withdraw at any time. No physical, psychological, or social risks were anticipated beyond those of ordinary computer use. The study protocol was reviewed and approved by the Institutional Review Board (IRB).
\input{fig/ap/nft}

%% file: algo/algo.tex
\begin{algorithm}[!t]
    \caption{\label{alg:distillation}\textbf{GDMD Training Procedure}}
    \KwIn{Pretrained teacher diffusion model $\mu_\text{real}$, online reward model $R_\psi$, 
    paired dataset $\mathcal{D}=\{z_\text{ref}, y_\text{ref}\}$,  hyperparameters $\lambda, \gamma, \eta$, discrete scheduler function $f_{dis}$. 
    }
    \KwOut{Trained model $G_\theta$.}
    
    \tcp{Initialize model}
    $G_\theta \leftarrow \text{copyWeights}(\mu_\text{real}),$
    $\mu_\text{fake}^{\phi} \leftarrow \text{copyWeights}(\mu_\text{real})$
    
    \While{train}{
        \tcp{\textbf{\textcolor{gray}{Prepare Data}}}
        Sample batch $z \sim \mathcal{N}(0, \mathbf{I})$ and $(z_\text{ref}, y_\text{ref}) \sim \mathcal{D}$, timestep $t$ from $f_\text{dis}$
        
        $z_t \leftarrow (1 - t)z + tz_\text{ref}$ \textbf{if} use data \textbf{else} BackSim($z,t,y_\text{ref}$)
        
        $x_0 \leftarrow \mathcal{S}(G_\theta(z_t,t,y_\text{ref}))$
        
        \text{~}

        \tcp{\textbf{\textcolor{gray}{Update fake score estimation model}}}
        \For{ i in range($\eta$)}
        {
            $\mathcal{L}_\text{Diff} \leftarrow \text{DenoisingLoss}(\mu_\text{fake}(x_t, t), stopgrad(x_0))$
            \hfill \tcp{Eq~\ref{eq:dmd-fake}}

            $\mu^\phi_\text{fake} \leftarrow \text{update}(\mu^{\phi}_\text{fake}, \mathcal{L}_\text{Diff})$

            $grads[i] \leftarrow \text{DistributionMatching}(\mu_\text{real}, \mu_\text{fake}^\phi, x_0)$
            \hfill \tcp{\textbf{\textcolor{blue}{DaGC}}, Eq~\ref{eq:dmd-gen}}
            
            $x_\text{tar}[i]  \leftarrow x_0 - grads[i] $ 
            \hfill \tcp{ Eq~\ref{eq:gradient-scoring}}

            $r^{raw}[i] \leftarrow R_\psi(\mathcal{D}_\text{vae}(x_\text{tar}[i]), y_\text{ref})$
            \hfill \tcp{ \textbf{\textcolor{blue}{IGS}}, Eq~\ref{eq:rraw}}
            
        }

        $r \leftarrow \text{GroupComputation}(r^{raw})$
        \hfill \tcp{Eq~\ref{eq:norm}}

        \text{~}
        
        \tcp{\textbf{\textcolor{gray}{Update generator}}}
        \For{i in range($\eta$)}
        {
            $\mathcal{L}_\text{DMD} \leftarrow \text{MSELoss}(x_0, stopgrad(x_\text{tar}[i])) $

            $\mathcal{L}_\text{RL} \leftarrow \text{DiffusionNFTLoss}(x_0, x_\text{tar}[i], r) $
            \hfill \tcp{\textbf{\textcolor{blue}{NaPO}},  Eq~\ref{eq:nft}}

            $G_\theta \leftarrow \text{update}(G_\theta, \lambda \mathcal{L}_\text{DMD} + \gamma \mathcal{L}_\text{RL})$
            
        }
    }
   \vspace{-0.2em}
\end{algorithm}

%% file: fig/ap/nft.tex
\begin{figure}[!t]
  \centering
  \setlength{\abovecaptionskip}{0cm}
  \begin{subfigure}{0.48\linewidth}
    \includegraphics[width=\linewidth]{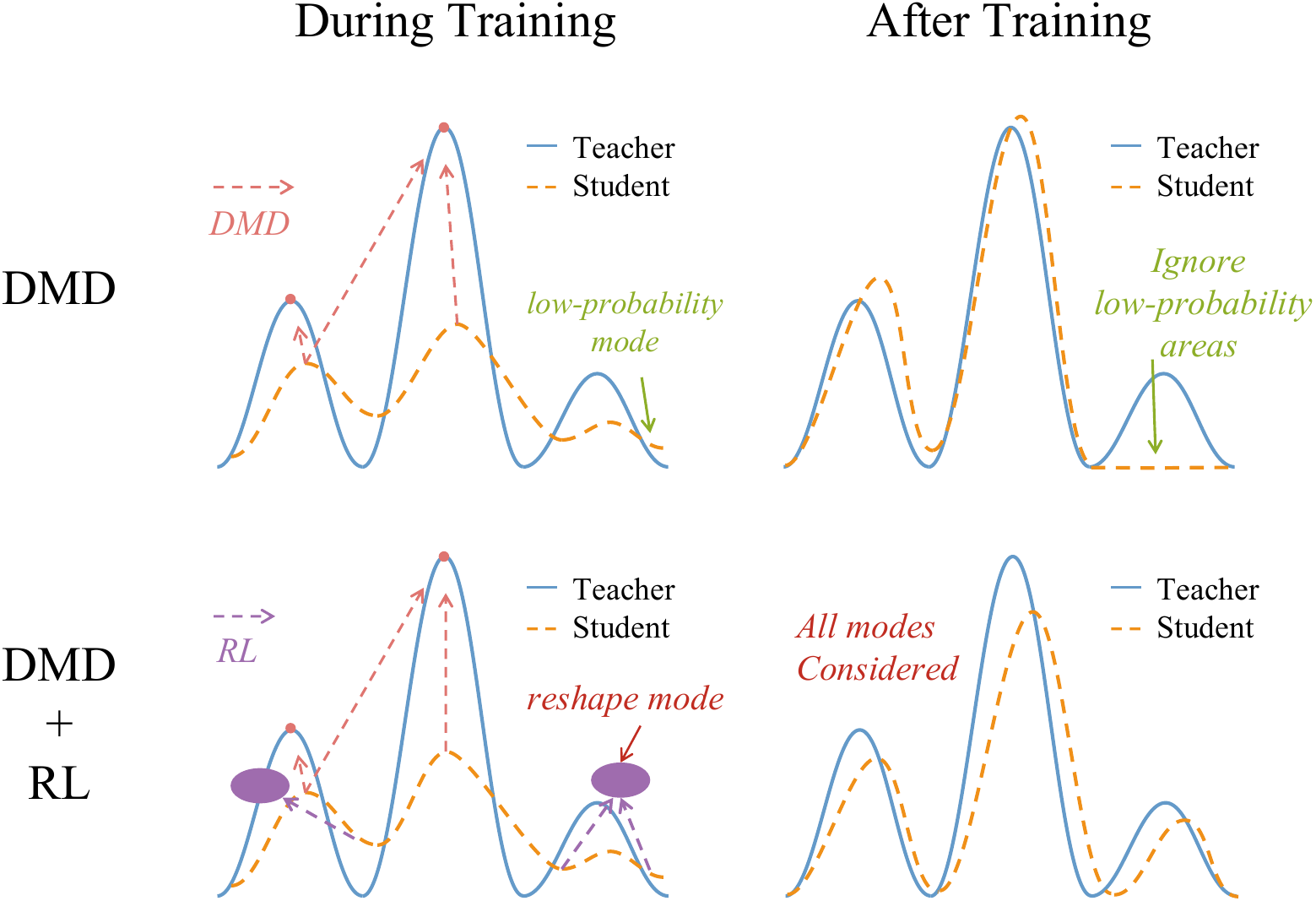}
    \caption*{}
  \end{subfigure}
  \hfill
    \begin{subfigure}{0.48\linewidth }
    \includegraphics[width=\linewidth]{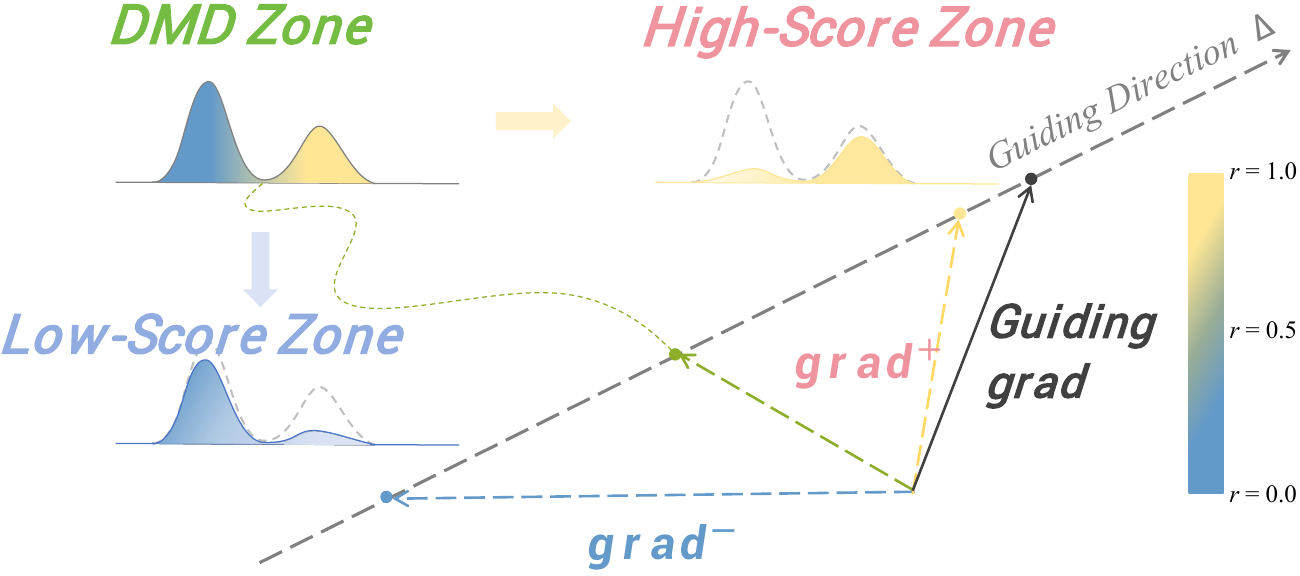}
    \caption*{}
  \end{subfigure}
  \caption{\textbf{Left:} Illustration of zero-forcing in DMD updates. 
  \textbf{Right:} Illustration of the optimization strategy based on implicit positive and negative gradient perception.}
  \label{fig:ap:mode}
\end{figure}

%% file: section/appendix/C_results.tex
\section{Additional Quantitative Results}
\label{sec:ap:results}
\label{sec:ap:rebuttal}

\subsection{Sampling Variability}
\label{sec:ap:seeds}
We repeat generation with ten randomly selected seeds and 5,000 images per seed. \Cref{tab:rebuttal:seeds} reports the mean and standard deviation across these ten runs for DMD2, DMDR, and GDMD on SDXL and SD3-Medium. GDMD has the highest mean on all five measures within each backbone. These statistics describe generation-seed variability.
\input{tables/rebuttal/seeds}


\subsection{Training Cost}
\label{sec:ap:cost}
\Cref{tab:rebuttal:cost} reports a separate profiling experiment on eight H100 GPUs. Collecting multiple gradient candidates increases iteration time from 0.715 to 1.265 seconds, or about $1.77\times$ (77\% overhead).
Thus, avoiding SDE rollouts does not make GDMD cheaper per iteration than DMDR. 
\input{tables/rebuttal/cost}

%% file: tables/rebuttal/seeds.tex
\begin{table}[t]
\centering
\caption{\textbf{Sampling variability over ten generation seeds.}
Each seed generates 5,000 images. Values are mean $\pm$ standard
deviation; these are not independent retraining runs.}
\label{tab:rebuttal:seeds}
\resizebox{\linewidth}{!}{%
\begin{tabular}{l c c c c c}
\toprule
\multirow{2}{*}{\textbf{Method}}
& \multicolumn{2}{c}{\textbf{Trained Rewards}}
& \multicolumn{3}{c}{\textbf{Unseen Rewards}} \\
\cmidrule(lr){2-3} \cmidrule(lr){4-6}
& CLIP Score$\uparrow$
& HP Score$\uparrow$
& Aesthetic Score$\uparrow$
& Pick Score$\uparrow$
& ImageReward$\uparrow$ \\
\midrule
DMD2-SDXL
& $0.2901\pm0.0004$
& $0.2936\pm0.0010$
& $5.7206\pm0.023$
& $22.4147\pm0.04$
& $0.8978\pm0.013$ \\
DMDR-SDXL
& $0.2900\pm0.0001$
& $0.2971\pm0.0006$
& $5.7903\pm0.017$
& $22.2170\pm0.04$
& $0.7531\pm0.008$ \\
\rowcolor[RGB]{240,230,245}
GDMD-SDXL
& $\mathbf{0.2912}\pm0.0002$
& $\mathbf{0.2993}\pm0.0009$
& $\mathbf{5.8107}\pm0.014$
& $\mathbf{22.5616}\pm0.05$
& $\mathbf{0.9185}\pm0.014$ \\
\arrayrulecolor{black!40}\midrule
DMD2-SD3
& $0.2908\pm0.0007$
& $0.2940\pm0.0016$
& $5.7679\pm0.011$
& $22.1130\pm0.06$
& $1.1497\pm0.021$ \\
DMDR-SD3
& $0.2901\pm0.0002$
& $0.2937\pm0.0007$
& $5.5201\pm0.009$
& $21.6388\pm0.06$
& $1.0068\pm0.024$ \\
\rowcolor[RGB]{240,230,245}
GDMD-SD3
& $\mathbf{0.2932}\pm0.0002$
& $\mathbf{0.3081}\pm0.0008$
& $\mathbf{5.8646}\pm0.009$
& $\mathbf{22.4979}\pm0.05$
& $\mathbf{1.2899}\pm0.019$ \\
\arrayrulecolor{black}\bottomrule
\end{tabular}%
}
\end{table}

%% file: tables/rebuttal/cost.tex
\begin{table}[htbp]
\centering
\caption{Training overhead measured on eight H100 GPUs, with batch size 2, group size 8, and image size 512.}
\label{tab:rebuttal:cost}
\small
\setlength{\tabcolsep}{4pt}
\begin{tabular}{lrrr}
\toprule
Method & iter/s & s/iter & TFLOPs/GPU \\
\midrule
DMDR & 1.398 & 0.715 & 23.3589 \\
GDMD & 0.791 & 1.265 & 41.2776 \\
\bottomrule
\end{tabular}
\end{table}

%% file: section/appendix/D_ablations.tex
\section{More Ablation Studies}
\label{sec:ap:ablation}
\input{tables/ap/ab_mul}
\subsection{Reward Normalization and Policy Updates}
\Cref{tab:ap} presents an ablation comparison of our additional strategies.
We observe that replacing either the score normalization or the original strategy does not significantly degrade the overall optimization process.
Our results indicate that Mean-Variance Normalization yields better unseen metrics, while soft update performs on par with the final approach.


\subsection{Balance Between Distillation and Reward Alignment}
\label{sec:ap:gamma}
We fix $\lambda=1$ and vary $\gamma$ in \cref{tab:rebuttal:gamma}. The default $\gamma=0.1$ balances reward improvement and visual quality; settings from 0.05 to 0.5 give useful gains in this experiment.
A smaller value (0.01) provides limited improvement and reduces HPS relative to DMD.
Larger values (1 or 2) can yield high Aesthetic Scores despite visibly oversaturated outputs, illustrating that an unseen automatic reward is not an independent guarantee of perceptual quality.
\input{tables/rebuttal/gamma}

\subsection{Separating Loss Weighting from Target Scaling}
\label{sec:ap:scaling}

Let $g$ denote a detached DMD score-difference direction, $w$ its detached adaptive weight, $r$ a detached preference weight, and $\bar{x}$ the detached current output. We consider two placements of the adaptive weight in the positive loss:
\begin{align}
\mathcal{L}_{\mathrm{inside}}
&= r \left\|x-\operatorname{sg}(\bar{x}-wg)\right\|_2^2,
\nonumber\\
\mathcal{L}_{\mathrm{outside}}
&= rw \left\|x-\operatorname{sg}(\bar{x}-g)\right\|_2^2.
\label{eq:scaling-placement}
\end{align}
At $x=\bar{x}$, these formulations yield identical gradients for fixed $r$, $w$, and $g$:
\begin{equation}
\left.\nabla_x \mathcal{L}_{\mathrm{inside}}\right|_{x=\bar{x}}
=
\left.\nabla_x \mathcal{L}_{\mathrm{outside}}\right|_{x=\bar{x}}
=
2rwg.
\label{eq:scaling-gradient}
\end{equation}
Thus, moving $w$ from the target displacement to the loss weight preserves the underlying distillation gradient at the current reference output.
The two formulations nevertheless construct different implicit targets:
\begin{align}
x_{\mathrm{tar}}^{\mathrm{inside}}
&= \operatorname{sg}(\bar{x}-wg),
\nonumber\\
x_{\mathrm{tar}}^{\mathrm{outside}}
&= \operatorname{sg}(\bar{x}-g).
\label{eq:scaling-targets}
\end{align}
Since the reward model evaluates the VAE-decoded target, this difference changes the image presented for reward evaluation.
In particular, placing $w$ inside the target directly scales its displacement from $\bar{x}$, whereas placing $w$ outside the target retains its role in weighting the distillation update without applying that scaling to the reward-model input. 

\paragraph{Placement of the adaptive weight.}
We apply the DMD-style sample-dependent weight defined in \cref{sec:ap:imple} to the loss (outside) rather than to the implicit target (inside).
\Cref{tab:rebuttal:scaling} compares these two placements and evaluates alternatives that replace the adaptive weight with the timestep $t$.
Moving the weight outside the target preserves the underlying distillation gradient at the current output while avoiding this amplification of the displacement in the image presented to the reward model.
The $t$-based alternatives exhibit less oversaturation but achieve lower scores than the DMD-style loss weighting.
We therefore retain the adaptive DMD weight as a loss coefficient.
\input{tables/rebuttal/scaling}

\paragraph{Magnitude of the target update.}
To examine the effect of target displacement independently of the adaptive loss weight, we introduce a separate coefficient $w_2$:
\begin{align}
x_{\mathrm{tar}}(w_2)
&= \operatorname{sg}(\bar{x}-w_2g),
\nonumber\\
\mathcal{L}_{\mathrm{pos}}
&= rw\left\|x-x_{\mathrm{tar}}(w_2)\right\|_2^2.
\label{eq:w2-target}
\end{align}
Here, $w$ remains outside the target, while $w_2$ controls the magnitude of the displacement used to construct the reward-model input.
Unlike relocating $w$, changing $w_2$ also scales the local loss gradient, which equals $2rww_2g$ at $x=\bar{x}$ for fixed $r$, $w$, and $g$.
Our default setting is $w_2=1$.
As shown in \cref{tab:rebuttal:w2}, reducing $w_2$ to 0.8 alleviates oversaturation while retaining improvements over the DMD baseline across all five evaluation metrics.
Further reductions diminish these gains; for example, at $w_2=0.5$, HPS and ImageReward fall below the DMD baseline, possibly because reward models struggle to distinguish output quality when the perturbation $w_2g$ is very small.
These results indicate that target scaling provides a practical means of balancing reward improvement against oversaturation, although excessive attenuation can compromise generation quality.
\input{tables/rebuttal/w2}



%% file: tables/ap/ab_mul.tex
\begin{table}[!t]
\centering
\captionof{table}{\textbf{Ablation study on score normalization and old policy designs}}
\resizebox{1\linewidth}{!}{
\begin{tabular}{l c c c c c}
\toprule
\multirow{2}{*}{\textbf{Method}} & \multicolumn{2}{c}{\textbf{Trained Rewards}} & \multicolumn{3}{c}{\textbf{Unseen Rewards}} \\
\cmidrule(lr){2-3} \cmidrule(lr){4-6}
& ~CLIP Score$\uparrow$ & ~HP Score$\uparrow$ & ~Aesthetic Score$\uparrow$ &~Pick Score$\uparrow$ & ~ImageReward$\uparrow$\\
\midrule
DMD Baseline & 0.2861 & 0.2891 & 5.5598 & 21.6216 & 0.9704 \\
\arrayrulecolor{black!40}\midrule
GDMD w/ Mean-Variance Normalization  & 0.2889 & 0.3048 & 5.8721 & \textbf{22.5004} & \textbf{1.2800} \\
GDMD w/ Soft Update & \textbf{0.2930} &  0.2964& 5.7854 & 22.3874 & 1.2329 \\
\rowcolor[RGB]{240,230,245}
GDMD & \textbf{0.2930} & \textbf{0.3076} & \textbf{5.8728} & {22.4614} & {1.2702} \\
\arrayrulecolor{black}\bottomrule
\end{tabular}}
\label{tab:ap}
\end{table}

%% file: tables/rebuttal/gamma.tex
\begin{table}[t]
\centering
\caption{\textbf{RL-weight ablation with $\lambda=1$.}
The default is $\gamma=0.1$.}
\label{tab:rebuttal:gamma}
\resizebox{\linewidth}{!}{%
\begin{tabular}{l c c c c c}
\toprule
\multirow{2}{*}{$\boldsymbol{\gamma}$}
& \multicolumn{2}{c}{\textbf{Trained Rewards}}
& \multicolumn{3}{c}{\textbf{Unseen Rewards}} \\
\cmidrule(lr){2-3} \cmidrule(lr){4-6}
& CLIP Score$\uparrow$
& HP Score$\uparrow$
& Aesthetic Score$\uparrow$
& Pick Score$\uparrow$
& ImageReward$\uparrow$ \\
\midrule
DMD baseline
& 0.2861 & 0.2891 & 5.5598 & 21.6216 & 0.9704 \\
\arrayrulecolor{black!40}\midrule
0.01
& 0.2915 & 0.2811 & 5.6026 & 21.7217 & 1.0338 \\
0.05
& 0.2921 & 0.2982 & 5.8127 & \textbf{22.4649} & 1.2651 \\
\rowcolor[RGB]{240,230,245}
0.1
& \textbf{0.2930} & 0.3076 & 5.8728 & 22.4614 & \textbf{1.2702} \\
0.5
& 0.2905 & \textbf{0.3082} & 5.8696 & 22.2395 & 1.2198 \\
1.0
& 0.2907 & 0.2998 & 5.9630 & 22.0221 & 1.1101 \\
2.0
& 0.2873 & 0.3037 & \textbf{6.0296} & 21.8546 & 1.1516 \\
\arrayrulecolor{black}\bottomrule
\end{tabular}%
}
\end{table}

%% file: tables/rebuttal/scaling.tex
\begin{table}[t]
\centering
\caption{\textbf{Effect of scaling placement.}
The scale is placed inside the implicit target or outside it as a loss
weight. The default uses the DMD-style loss weight (outside).}
\label{tab:rebuttal:scaling}
\resizebox{\linewidth}{!}{%
\begin{tabular}{l c c c c c}
\toprule
\multirow{2}{*}{\textbf{Method}}
& \multicolumn{2}{c}{\textbf{Trained Rewards}}
& \multicolumn{3}{c}{\textbf{Unseen Rewards}} \\
\cmidrule(lr){2-3} \cmidrule(lr){4-6}
& CLIP Score$\uparrow$
& HP Score$\uparrow$
& Aesthetic Score$\uparrow$
& Pick Score$\uparrow$
& ImageReward$\uparrow$ \\
\midrule
DMD baseline
& 0.2861 & 0.2891 & 5.5598 & 21.6216 & 0.9704 \\
\arrayrulecolor{black!40}\midrule
DMD-style (inside)
& 0.2760 & 0.2513 & 5.4142 & 20.4131 & -0.7928 \\
\rowcolor[RGB]{240,230,245}
DMD-style (outside)
& \textbf{0.2930}
& \textbf{0.3076}
& \textbf{5.8728}
& \textbf{22.4614}
& \textbf{1.2702} \\
$t$-style (inside)
& 0.2894 & 0.2707 & 5.6941 & 21.6657 & 0.9008 \\
$t$-style (outside)
& 0.2836 & 0.2755 & 5.7614 & 21.6424 & 0.8133 \\
\arrayrulecolor{black}\bottomrule
\end{tabular}%
}
\end{table}

%% file: tables/rebuttal/w2.tex
\begin{table}[t]
\centering
\caption{\textbf{Target-update scale ablation.}}
\label{tab:rebuttal:w2}
\resizebox{\linewidth}{!}{%
\begin{tabular}{l c c c c c}
\toprule
\multirow{2}{*}{$\boldsymbol{w_2}$}
& \multicolumn{2}{c}{\textbf{Trained Rewards}}
& \multicolumn{3}{c}{\textbf{Unseen Rewards}} \\
\cmidrule(lr){2-3} \cmidrule(lr){4-6}
& CLIP Score$\uparrow$
& HP Score$\uparrow$
& Aesthetic Score$\uparrow$
& Pick Score$\uparrow$
& ImageReward$\uparrow$ \\
\midrule
DMD baseline
& 0.2861 & 0.2891 & 5.5598 & 21.6216 & 0.9704 \\
\arrayrulecolor{black!40}\midrule
\rowcolor[RGB]{240,230,245}
1.0 (default)
& \textbf{0.2930}
& \textbf{0.3076}
& \textbf{5.8728}
& \textbf{22.4614}
& \textbf{1.2702} \\
0.8
& 0.2898 & 0.3014 & 5.8385 & 22.3915 & 1.1831 \\
0.5
& 0.2881 & 0.2765 & 5.6879 & 21.9239 & 0.8193 \\
0.2
& 0.2839 & 0.2712 & 5.6268 & 21.7482 & 0.7591 \\
\arrayrulecolor{black}\bottomrule
\end{tabular}%
}
\end{table}

%% file: section/appendix/E_qualitative.tex
\clearpage
\section{Qualitative Results and Evaluation Prompts}
\label{sec:ap:vis}
\subsection{Generated Samples and Baseline Comparisons}
We provide additional results from GDMD 4-NFE sampling in \cref{fig:ap_vis}, followed by comparisons with existing methods in \cref{fig:ap_com}.
\input{fig/ap/ap_vis}
\subsection{Training Progress}
\label{sec:ap:progress}
\Cref{fig:ap_speed} illustrates how generated samples change during training. Quantitative computational costs are reported separately in \cref{sec:ap:cost}.
\clearpage
\input{fig/ap/ap_com}
\input{fig/ap/ap_speed}

\clearpage
\subsection{Text Prompts Used in Sampling}
\label{sec:ap:text}
Here we list the text prompts used in \cref{fig:teaser} and \cref{fig:ap_vis}.

In \cref{fig:teaser} (From top to bottom, from left to right):
\begin{itemize}
    \item ``From a slight low angle, an androgynous figure in a fedora, their long face sharply illuminated by cool, intense side lighting, asymmetrically surveys a Gigeresque landscape.''
    \item ``Nestled within Indonesia's verdant jungles lies the largest single flower in the world, flourishing in solitude.'' 
    \item ``A handful of tiny, nomadic yurts are scattered across an immense, windswept steppe under a vast, cloud-strewn sky.''
    \item ``A beagle wearing sunglasses, caught in profile by the cool, moody side glow of bioluminescent fungi.''
    \item ``On a warm, humid afternoon, a frosty glass of ice water sits on a wooden table, highlighting the droplets of condensation clinging to its surface.''
    \item ``A flame from a candle dancing lightly before slowly dimming within a sealed glass vessel.''
    \item ``A vibrant ice cream cone dripping with colorful trails and drooping sides, softened by the intense heat of the summer sun.''
    \item ``A dining table with a vase of colorful flowers in front of a white kitchen background.''
    \item ``An outdoor setting bathed in bright light, showcasing a dilapidated chair amid scattered trash, inspiring feelings of discomfort and revulsion.''
    \item ``A giraffe displaying a humorous facial expression.''
\end{itemize}

In \cref{fig:ap_vis} (From top to bottom, from left to right):
\begin{itemize}
    \item ``An imposing Italian wolf perched on a rugged hillside within a forest, illuminated by the warm glow of a golden sunset.''
    \item ``A candle drifting through the expanse of space, encircled by shimmering stars.''
    \item ``A gray and white cat sits near a car window.''
    \item ``A leaf shimmering with frost in the early morning light.''
    \item ``a wooden spatula with paint on it and a rubber band that has been stretched.''
    \item ``A charcoal-sketched Middle Eastern person's blissful, closed-eye smile is intricately revealed by cool rim light against a warm, crackling fireplace's glow, captured frontally in a golden spiral.''
    \item ``A pot filled with lightly salted water showing subtle ripples and a thin trail of steam, suggesting it is nearing but has not yet reached a full boil.''
    \item ``Create an image: A nighttime photograph depicts a serene forest clearing illuminated by a glowing lantern suspended from the branches of a centuries-old oak tree.''
    \item ``A few solitary, wind-battered trees cling precariously to the edge of a vast, rocky cliff overlooking a turbulent, grey sea.''
    \item ``A lone flower in full bloom resting among the grass, gently covered by the shadow cast from a branch overhead, demonstrating delicate occlusion.''
    \item ``Dynamic close-up: a business professional's face, mid-leap with scarf trailing, dramatically sculpted by cool top-down twilight against a warm, richly ornamented, blurred background.''
    \item ``A few colorful hot air balloons drift serenely over vast, mist-covered rolling hills at sunrise.''
    \item ``A hotel bed, perfectly arranged, featuring a soft blanket and a stack of clean towels placed in the middle.''
    \item ``An undersized teacup placed atop a large saucer adorned with a floral design.''
    \item ``A vast expanse of fractured ice floes drifts slowly across a dark, frigid Arctic ocean under a pale, ethereal aurora.''
    \item ``A vast flock of starlings creates dynamic, amorphous shapes against a soft, pastel-colored sunset sky over rolling, empty hills.''
    \item ``A stone beneath supports the earth, swaying on the edge of imbalance.''
    \item ``A freshly baked pizza topped with various ingredients sits on a plate alongside a shaker of seasoning on a dimly lit table.''
\end{itemize}

%% file: fig/ap/ap_vis.tex
\begin{figure}[!t]
  \centering
    \includegraphics[width=\linewidth]{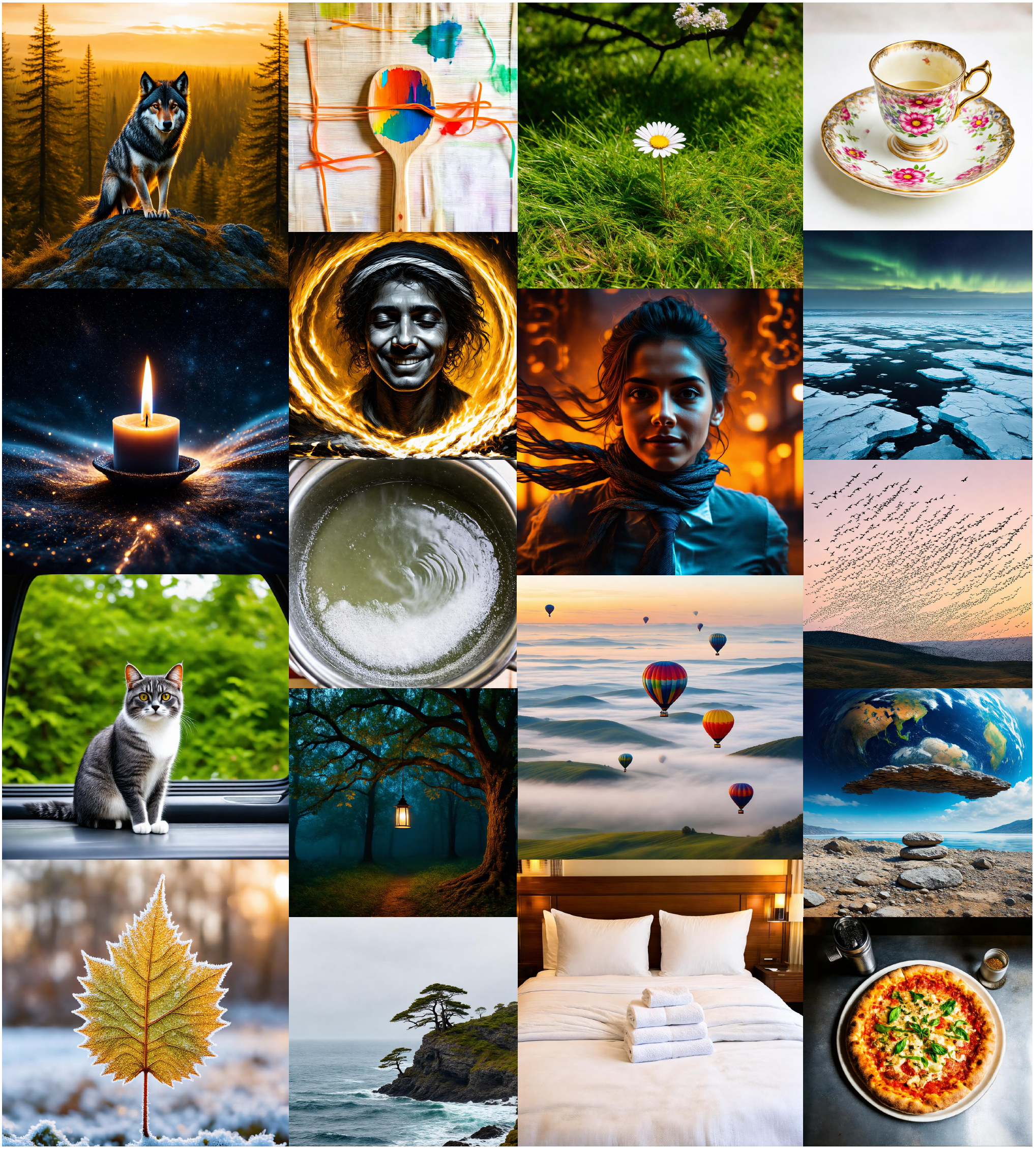}
   \caption{\textbf{Samples from 4-NFE student model distilled through our methods.}
   }
   \label{fig:ap_vis}
\end{figure}

%% file: fig/ap/ap_com.tex
\begin{figure}[!t]
  \centering
    \includegraphics[width=\linewidth]{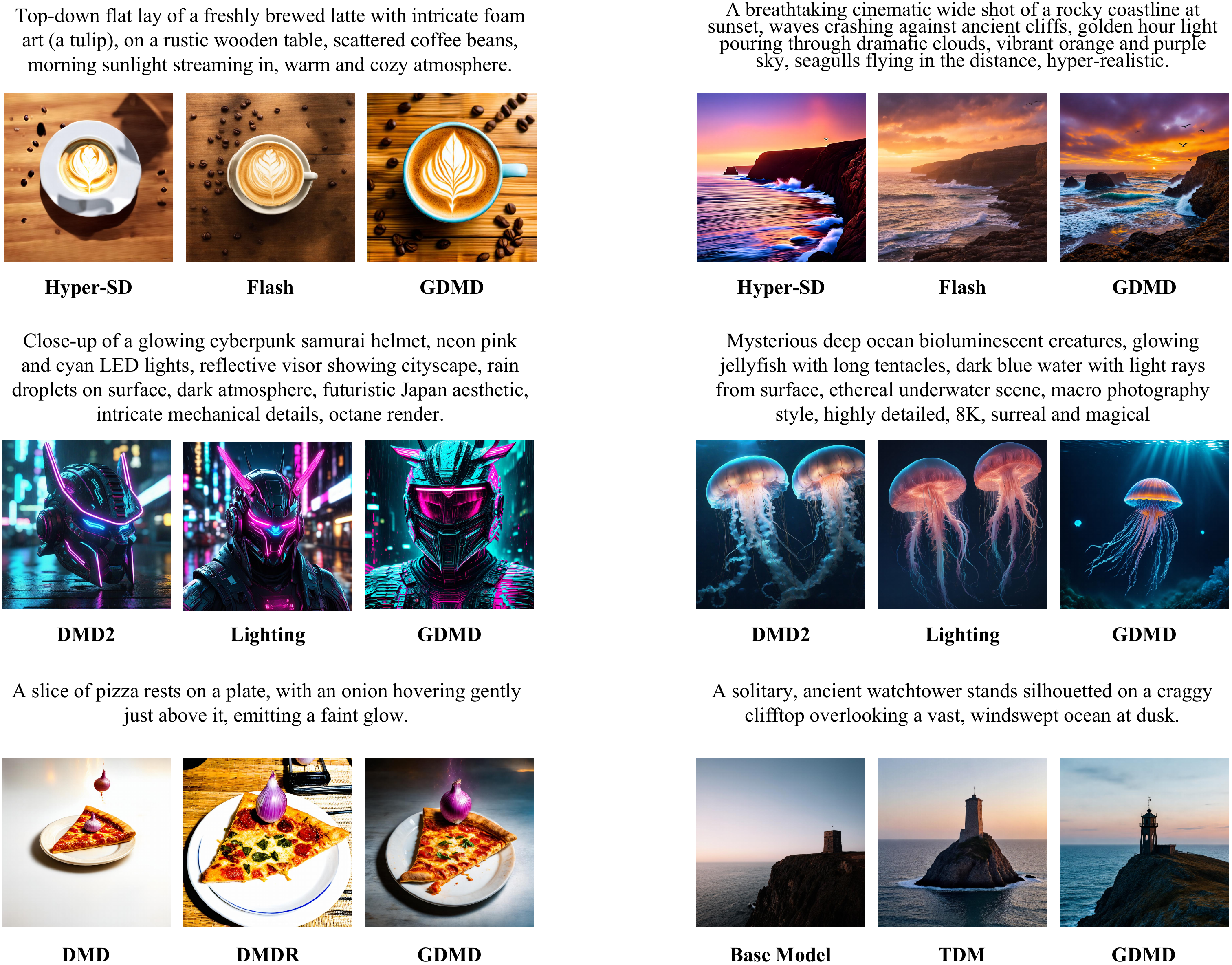}
   \caption{\textbf{Visual comparison with existing methods and the base model.}
   }
   \label{fig:ap_com}
   \vspace{-0.2em}
\end{figure}

%% file: fig/ap/ap_speed.tex
\begin{figure}[!t]
  \centering
    \includegraphics[width=\linewidth]{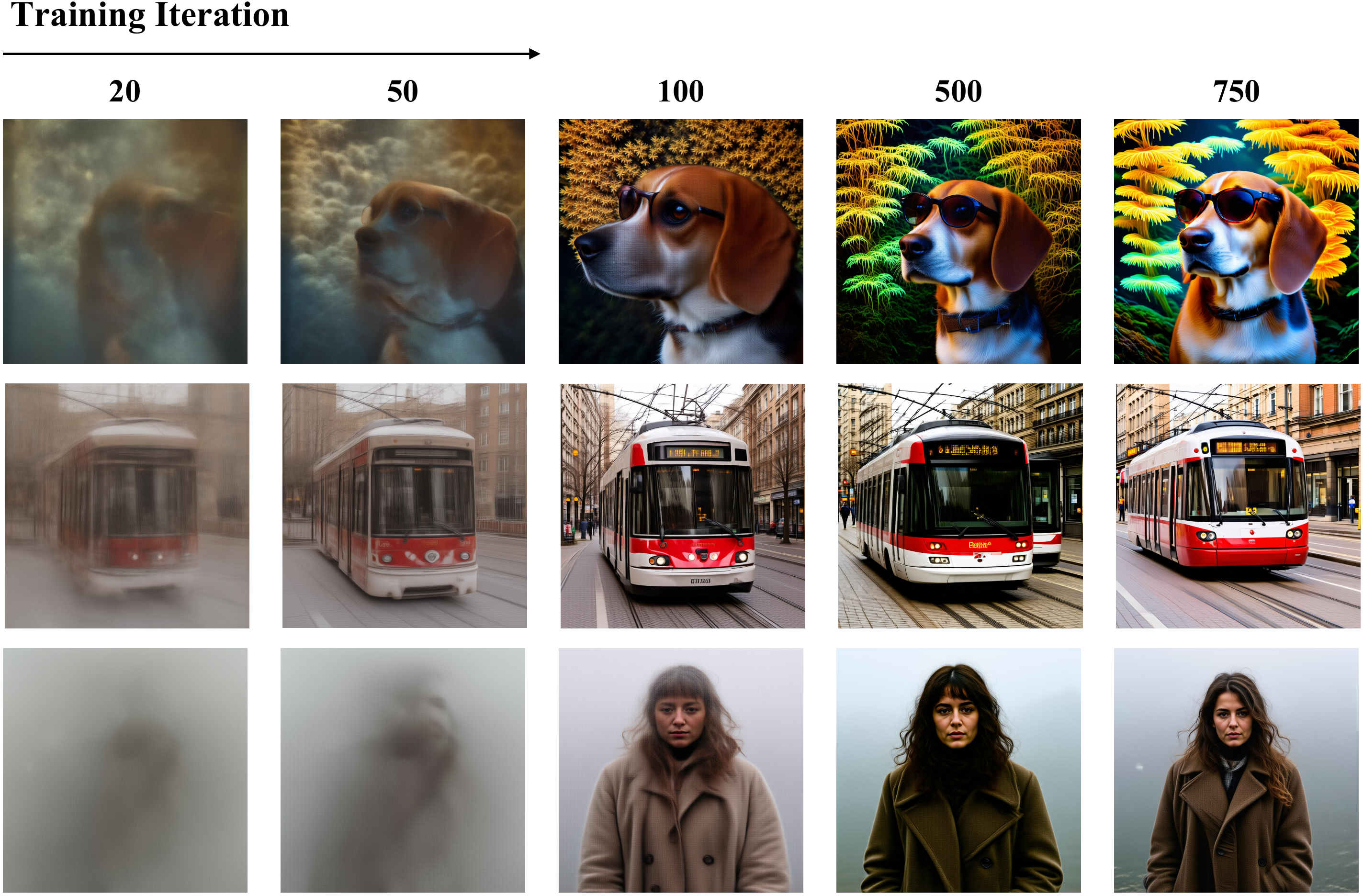}
   \caption{\textbf{Visualization of training efficiency.}
   }
   \label{fig:ap_speed}
\end{figure}

%% file: section/appendix/F_impact.tex
\section{Broader Impact}
\label{sec:ap:bro}
Generative models have made remarkable progress, now routinely producing realistic, high‑quality images, videos, and other modalities \cite{luo2024large, miao2025advances}. 
GDMD offers a universal solution to significantly accelerate sampling from diffusion and flow‑based generative models via distillation. 
Such real‑time generation enables interactive applications and supports creative expression, while also improving inference efficiency and reducing energy consumption.
That said, diffusion and flow models, including their distilled variants, can be misused to create deceptive content. 
Therefore, they must be deployed with due caution.
Additionally, we find that training with AIGC-synthesized data is more prone to push optimization toward oversaturation or collapse. In particular, jointly distilling a weaker model (e.g., SDXL) from synthetic data generated by a stronger model (e.g., FLUX) more easily leads to oversaturation and reward hacking. Another limitation is that RL-enhanced distillation may suffer from instruction failure or generation collapse under complex prompts. Because RL operates as a preference-alignment mechanism, it may tighten the diffusion distribution, potentially reducing robustness in complex or unseen scenarios and leading to meaningless outputs or severe reward hacking. This limitation suggests that extra caution is needed when deploying such methods in important real-world applications.